\documentclass[10pt,twocolumn,letterpaper]{article}

\usepackage{cvpr}              %

\usepackage{graphicx}
\usepackage{amsmath}
\usepackage{amssymb}
\usepackage{booktabs}
\usepackage{makecell}
\usepackage{multirow}
\usepackage{array}
\usepackage{comment}
\usepackage{xcolor}
\usepackage{soul}
\usepackage[skip=.6em]{caption}

\usepackage[pagebackref,breaklinks,colorlinks]{hyperref}

\usepackage[capitalize]{cleveref}
\crefname{section}{Sec.}{Secs.}
\Crefname{section}{Section}{Sections}
\Crefname{table}{Table}{Tables}
\crefname{table}{Tab.}{Tabs.}

\newcommand{\todo}[1]{}
\renewcommand{\todo}[1]{{\color{blue} {#1}}}

\def\cvprPaperID{11036} %
\def\confName{CVPR}
\def\confYear{2022}

\begin{document}

\title{P$^3$IV: Probabilistic Procedure Planning from Instructional Videos \\ with Weak Supervision}
\author{He Zhao$\mathbf{^{1,2}}$
\quad
Isma Hadji$\mathbf{^1}$
\quad
Nikita Dvornik$\mathbf{^1}$
\quad
Konstantinos G. Derpanis$\mathbf{^{1,2}}$\\
\quad
Richard P. Wildes$\mathbf{^{1,2}}$
\quad
Allan D. Jepson$\mathbf{^1}$\vspace{1pt}
\quad\\
{\large $\mathbf{^1}$Samsung AI Centre Toronto, $\mathbf{^2}$York University} \\
{\tt\small \{zhufl, kosta, wildes\}@eecs.yorku.ca},
{\tt\small \{isma.hadji, n.dvornik, allan.jepson\}@samsung.com}
}
\maketitle

\begin{abstract}

In this paper, we study the problem of procedure planning in instructional videos. %
Here, an agent must produce a plausible sequence of actions that can transform the environment from a given start to a desired goal state. 
When learning procedure planning from instructional videos, most recent work leverages intermediate visual observations as supervision, which requires expensive annotation efforts to localize precisely all the instructional steps in training videos.
In %
contrast, we remove the need for expensive temporal video annotations and propose a weakly supervised approach by learning from natural language instructions. %
Our model %
is based on a transformer equipped with a memory module, which maps the start and goal observations to a sequence of plausible actions.
Furthermore, we augment our model with a probabilistic generative module to capture the uncertainty inherent to procedure planning, an aspect largely overlooked by previous work.
We evaluate our model on three %
datasets and show our weakly-supervised approach %
outperforms previous fully supervised state-of-the-art models %
on multiple metrics.
\end{abstract}
\vspace{-9pt}
\section{Introduction}

\begin{figure}[t]
	\centering
	\scalebox{0.6}{ %
		\includegraphics[scale=0.3]{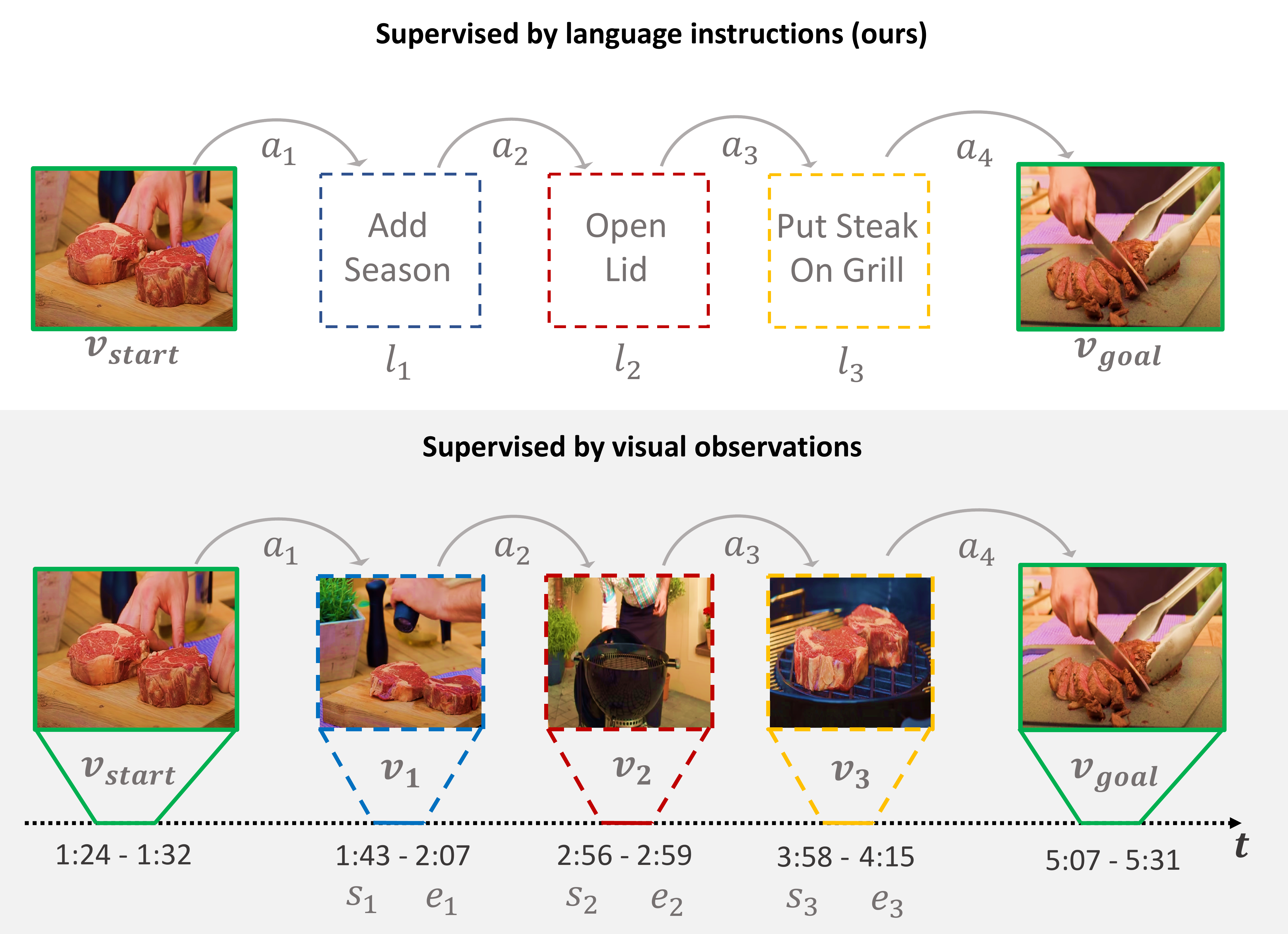}}
	\caption{
	Illustration of weak language supervision for procedure planning.
	Fully supervised approaches (bottom row) learn models from step labels, $a_{i}$, and intermediate visual representations, $v_{i}$, over $T$ finite steps. This strategy requires knowing the starting, $s_i$, and ending, $e_i$, timestamps, for each intermediate step. In contrast, our approach (top row) exploits natural language representations, $l_i$, of the intermediate labels, $a_{i},$ as a surrogate supervision, which only requires labeling the order of events. Note that the action label, $a_i$, is a discrete variable, whereas the action language representation, $l_i$, is a pre-trained continuous embedding.
	}
	\label{fig:teaser}
\end{figure}

Procedure planning is a natural task for humans -- one must plan out a sequence of actions that takes one from the current state to the desired goal. While effortless for humans, procedure planning is notoriously hard for artificial agents.
Nevertheless, solving procedure planning is of great importance for building next-level 
artificial intelligence
systems capable of analyzing and mimicking human behaviour, and eventually assisting humans in goal-directed problem solving, \textit{e.g.,} cooking, assembling furniture or tasks that can be represented as a clear set of instructions.
Traditionally, procedure planning has been addressed in structured environments, such as object manipulation on a table surface~\cite{finn2017deep,srinivas2018universal}.
While restricting the environment helps improve planning, it also limits the range of possible applications.
Here, we follow more recent work~\cite{procedure2020} and tackle procedure planning in the realm of instructional videos~\cite{COIN, CrossTask}. Given visual observations of the start and goal states, the task is to predict a sequence of high-level actions needed to achieve a goal; see Fig~\ref{fig:teaser}.
This task is particularly challenging as it requires parsing unstructured environments, recognizing human activities and understanding human-object interactions.
Yet, the range of applications for such planners is broad, which motivates research efforts on this problem.

Current approaches for procedure planning from instructional videos share a serious limitation -- reliance on strong supervision with expensive annotations~\cite{procedure2020, bi2021procedure, sun2021plate}.
Specifically, all such methods require access to (i) a list of action labels used to transition from start to the goaafl state and (ii) the visual representation of the intermediate states.
Using such intermediate visual representations entails 
very expensive annotation of the start and end times of each intermediate instructional step; see Fig~\ref{fig:teaser} (bottom).
In contrast, our work removes the need for intermediate visual states during training and instead uses their linguistic representation for supervision.
Relying on language representations allows us to better leverage instructional videos and significantly reduce the labeling effort; see Fig.~\ref{fig:teaser} (top).
For example, language annotations for the intermediate instructional steps could be extracted from %
general procedure descriptions available in recipes or websites, \textit{e.g.,} WikiHow \cite{wikihow}.
In contrast, to obtain the timestamps for intermediate instructional steps one must watch the entire instructional video.
In addition, a language representation can be a more stable supervisory signal~\cite{radford2021learning}, as the description of a given step (\eg, \textit{add seasoning}) remains the same, while its visual observation varies across different videos.

Previous work on procedure planning from video relies on a two-branch autoregressive approach while adopting different architectures to model these branches~\cite{procedure2020, bi2021procedure, sun2021plate}. In such models, one branch is dedicated to predicting actions based on the previous observation, while the other approximates the observation given the previous action in a step-by-step manner. 
Such models are cumbersome and 
compound errors, 
especially for longer sequences.
In contrast, we rely on a %
single branch non-autoregressive model,
implemented as a transformer \cite{vaswani2017attention} that
generates all intermediate steps in parallel 
conditioned on the start and goal observations.

Another %
important factor in procedure planning is to model the uncertainty %
inherent to the prediction task. %
For example, given a set of ingredients and the goal of making a pancake, the intermediate steps could be either (i) [\textit{add wet ingredients} $\xrightarrow{}$ \textit{add dry ingredients} $\xrightarrow{}$ \textit{whisk mixture}] or (ii) [\textit{add dry ingredients} $\xrightarrow{}$ \textit{add wet ingredients} $\xrightarrow{}$ \textit{whisk mixture}]. This example shows that in realistic scenarios some plans can vary even under a shared common goal.
This observation is usually handled in physical path planning tasks (\eg, robotic arms are  allowed to follow multiple feasible trajectories \cite{srinivas2018universal});
yet, effort is lacking on probabilistic modeling of procedure planning from instructional videos. While previous work included a probabilistic component at training time \cite{bi2021procedure}, 
we are the first to use and benefit from multiple plausible plans at inference.
We explicitly handle uncertainty in procedure planning with a dedicated generative module that can produce multiple feasible plans.

\textbf{Contributions.} In summary, the main technical contributions of our work are threefold. (i) We introduce a weakly supervised approach for procedure planning, which leverages powerful language representations extracted from pre-trained text-video embeddings. (ii) We tackle the task with a simpler single branch model, which can generate all intermediate steps in parallel, rather than relying on the two-branch auto-regressive approach used in previous work. %
(iii) We propose a generative adversarial framework, trained with an extra adversarial objective, to capture the stochastic property of planned procedures.
We evaluate our approach on three widely used instructional videos datasets and show state-of-the-art performance across different prediction time horizons, even while relying on weaker supervision. We also show %
the advantage of modeling uncertainty.
Our code is  available at: \url{https://github.com/SamsungLabs/procedure-planning}.

\section{Related work}
\textbf{Procedure planning.} Traditionally, %
goal-conditioned planning has been studied mostly in physical environments, \eg, robotic motion planning \cite{kaelbling1993hierarchical, florensa2018automatic, ghosh2018learning} and human pedestrian trajectory planning \cite{mangalam2020not}.
Recently, the task of procedure planning from instructional videos was introduced \cite{procedure2020}. Various approaches have made use of recurrent neural networks (RNNs) \cite{procedure2020}, transformers \cite{sun2021plate} and adversarial policy planning \cite{bi2021procedure}; all have used two-branches and strong visual supervision.
In contrast, we model the actions directly using a non-autoregressive transformer-based architecture.
More importantly, 
we use low-budget weak supervision in the form of language instructions instead of supervising the model with ``costly'' visual observations as done by all existing approaches.

\vspace{5pt}
\textbf{Supervision with natural language.} A common alternative to training visual models using manually defined label sets is to exploit semantic supervision from natural language.
Using natural language as supervision has several advantages: (i) language annotations can be collected automatically~\cite{miech19howto100m}; (ii) modeling language and vision jointly can produce stronger representations~\cite{desai2021virtex}; (iii) such supervision can achieve better generalization to unseen domains~\cite{radford2021learning}.
Such benefits resulted in a growing interest in using language as supervision for variety of tasks, \eg, image classification \cite{gomez2017self, joulin2016learning, desai2021virtex}, representation learning~\cite{miech2020end, videoclip}, video retrieval~\cite{miech19howto100m, gabeur2020multi}, step localization \cite{drop-dtw, miech2020end} and navigation with instruction following~\cite{pashevich2021episodic}.
We use advances in joint video and language modeling for procedure planning.
We use pre-trained features~\cite{miech2020end} to map language and video to a common space and replace expensive video supervision with readily available language instructions

\vspace{5pt}
\textbf{Sequence modeling with transformers.} Procedure planning is a task of conditional sequence prediction and thus it directly benefits from recent advances in sequence modeling. One of the strongest recent approaches to sequence modeling is the transformer architecture~\cite{vaswani2017attention}%
, which has been adopted for a wide variety of tasks, \eg, images~\cite{dosovitskiy2020image}, videos~\cite{arnab2021vivit} and multi-modal data~\cite{vilbert, Luo2020UniVL, videoclip} tasks. 
Recent work adopted the transformer decoder architecture for fixed-size set prediction 
via learnable input queries~\cite{carion2020end,zhang2021temporal}.
We build upon similar ideas by setting the first and last queries to correspond to the start and goal observations, while making the intermediate queries learnable. 
To improve long-range sequence modeling and help overall sequence coherence, recent work augmented transformers with an explicit external memory~\cite{lei2020mart,wu2020memformer}. 
For the same reasons, we also integrate a learnable memory module.%

\vspace{5pt}
\textbf{Future prediction.} %
The task of procedure planning is closely related to future prediction, 
where only past observations are provided as input.
A key consideration in future prediction is modeling prediction uncertainty from partial initial observations. 
One common approach for modeling uncertainty is a Variational Auto Encoder (VAE) \cite{vae} that captures the distribution over future actions.
Another approach is to use generative adversarial networks (GANs) to forecast multiple, distinct and high quality future activities \cite{zhao2020diverse, piergiovanni2020adversarial}. In this work, we adopt the generative modeling framework to model distributions over possible plans.

\section{Technical approach}
Here, %
we present our approach to procedure planning %
that relies on three main components.
First, we predict all steps in the plan in parallel using a non-autoregressive transformer decoder. To obtain coherent plan predictions, %
our transformer is augmented with a learned memory shared across all possible tasks in a given dataset (Sec.~\ref{sec:mem_trans}). Second, to model the uncertainty inherent to the task, we include a generative component trained with an adversarial loss. As a result, we can infer multiple feasible plans conditioned on start and goal observations (Sec.~\ref{sec:gan}).
Third, to supervise the transformer's outputs, we use the cross-entropy loss on action predictions and a contrastive loss to match visual state predictions with the corresponding language descriptions (Sec.~\ref{sec:losses}).
Fig.~\ref{fig:alg} provides an overview of our approach, which we detail next.%

\subsection{Problem formulation}
\begin{figure*}[t]
	\centering
    \includegraphics[trim=18 245 10 33,clip,width=0.99\textwidth]{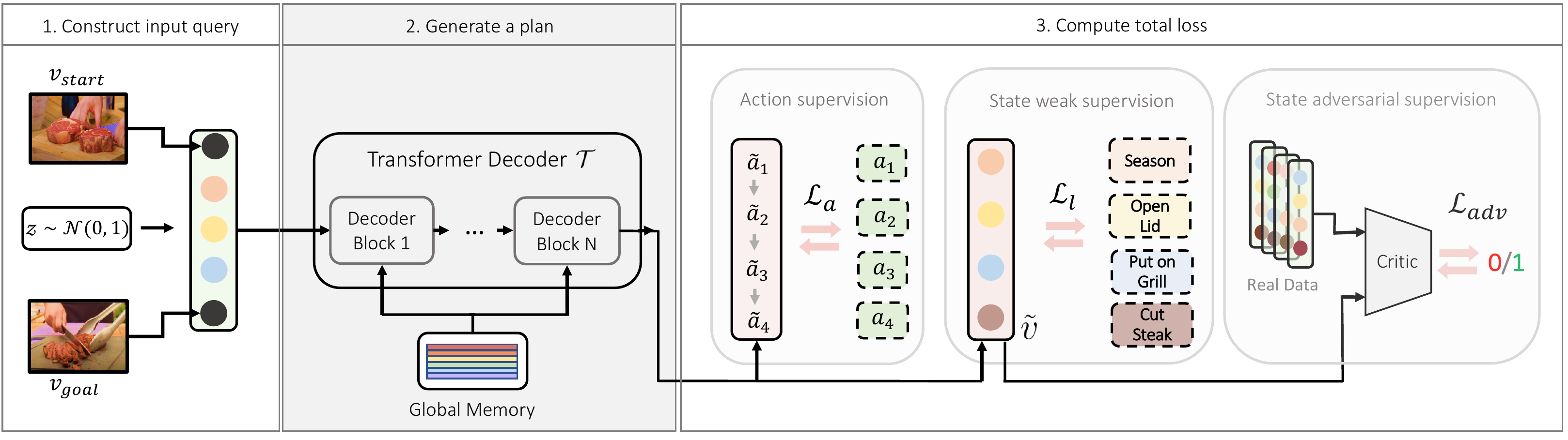}
	\caption{Overview of our procedure planning approach.
	First, we embed the visual observations of the start and goal states (\ie, black nodes), attach them to the sequence of learned queries (\ie, colored nodes) and add random noise to the resulting input sequence. Second, we pass the input to the transformer decoder that interacts with the global memory to generate feasible procedure plans. Third, we produce state and action vectors, $\{ \tilde{v}_i , \tilde{a}_i \}$,  and use a number of losses, $\mathcal{L}$ (described in Sec~\ref{sec:losses}), to supervise our architecture.
	}\label{fig:alg}
	\vspace{-10pt}
\end{figure*}

Given a start visual observation, $v_{start}$, and a desired visual goal, $v_{goal}$, our task is to predict a \emph{plan} defined as the sequence of $T$ intermediate action steps, $\tilde{\pi} = \tilde{a}_{1:T}$, taken to transition from $v_{start}$ to $v_{goal}$.
We overscore with $\sim$ to indicate our predictions, while the lack of such overscoring indicates ground truth (GT).
At training time, given $v_{start}$ and $v_{goal}$, we predict a plan, $\tilde{\pi}$, and corresponding visual observations, $\tilde{v}_{1:T}$.
We use the intermediate action labels, $a_{1:T}$, to train the plan prediction, $\tilde{\pi}$, and corresponding language descriptions (embedded with a pre-trained text encoder), $l_{1:T}$, to supervise the intermediate visual observations, $\tilde{v}_{1:T}$. %
That is, we substitute the visual information about intermediate instruction steps, $v_{1:T}$, with their language counterparts, $l_{1:T}$, to train the planner; see Fig.~\ref{fig:alg}.
We believe such supervision substitution is meaningful, as we are using a strong pre-trained vision-language encoder~\cite{miech2020end}, mapping visual activities and their descriptions in a common space, thus making visual, $v_{t}$, and language, $l_{t}$, features, corresponding to the same activity, interchangeable for training.
In contrast, previous work assumes access to the set of intermediate action-observation pairs (\ie, ${a_{1:T},v_{1:T}}$)~\cite{procedure2020, sun2021plate, bi2021procedure}, thereby requiring strong supervision to identify all intermediate visual observations. At inference time, we only use the start and goal observation, to predict a plan, $\tilde{\pi} = \tilde{a}_{1:T}$, for a given time horizon, $T$.

\subsection{Memory augmented transformer decoder}\label{sec:mem_trans}
To implement our planner, we use a non-autoregressive transformer decoder architecture \cite{carion2020end, zhang2021temporal}. Our transformer decoder takes two input types; namely, learnable queries augmented with the start and goal observations and a learned memory component, and outputs action and intermediate state predictions, as illustrated in Fig.~\ref{fig:alg}.\\

\noindent\textbf{Conditioned learned-query input.} The first input is the query set, $\mathbf{Q}=[q_{start}, q_1,\dotsc,q_{T-1}, q_{goal}]$, where the first and last inputs correspond to the representations of our initial and goal visual observations, $v_{start}$ and $v_{goal}$, respectively, while $q_{1:T-1}$ are a set of learned queries. Queries, $[q_1,\dotsc, q_{goal}]$, are associated with the action labels, $a_{1:T}$, that we wish to predict.
To communicate information about the order of elements to the decoder, we add to each query a fixed cosine positional embedding~\cite{bello2019attention}, $p_t$, as follows
\begin{equation}
\mathbf{Q} = [q_{start} + p_0,\dotsc, q_t + p_t,\dotsc, q_{goal} + p_T], 
\label{eq:input}
\end{equation}
\noindent where $q_t$ and $p_t$ all are encoded as $d$ dimensional embeddings, (\ie, $q_t$,  $p_t$ $\in \mathbb{R}^{d}$) and $t = 1,\dotsc, T-1$. \\

\noindent\textbf{Learned memory input.} The second input to our transformer decoder is a learned memory component that is common across all examples in a given dataset. The memory is defined as a set of $d$-dimensional vectors
\begin{equation}
    \textbf{M} = [m_1, m_2,\dotsc, m_n] \in \mathbb{R}^{d \times n},
\label{eq:mem}
\end{equation}
where $n$ is the number of learnable vectors in the memory bank. Notably, the size of the memory (\ie, number of entries, $n$, in the memory) is a hyperparameter that is independent from the prediction time horizon. We  %
use read-only memory \cite{marchetti2020mantra} and share it among all layers for simplicity.

\begin{figure}[t]
\centering
\includegraphics[trim=180 50 260 50,clip,width=0.75\columnwidth]{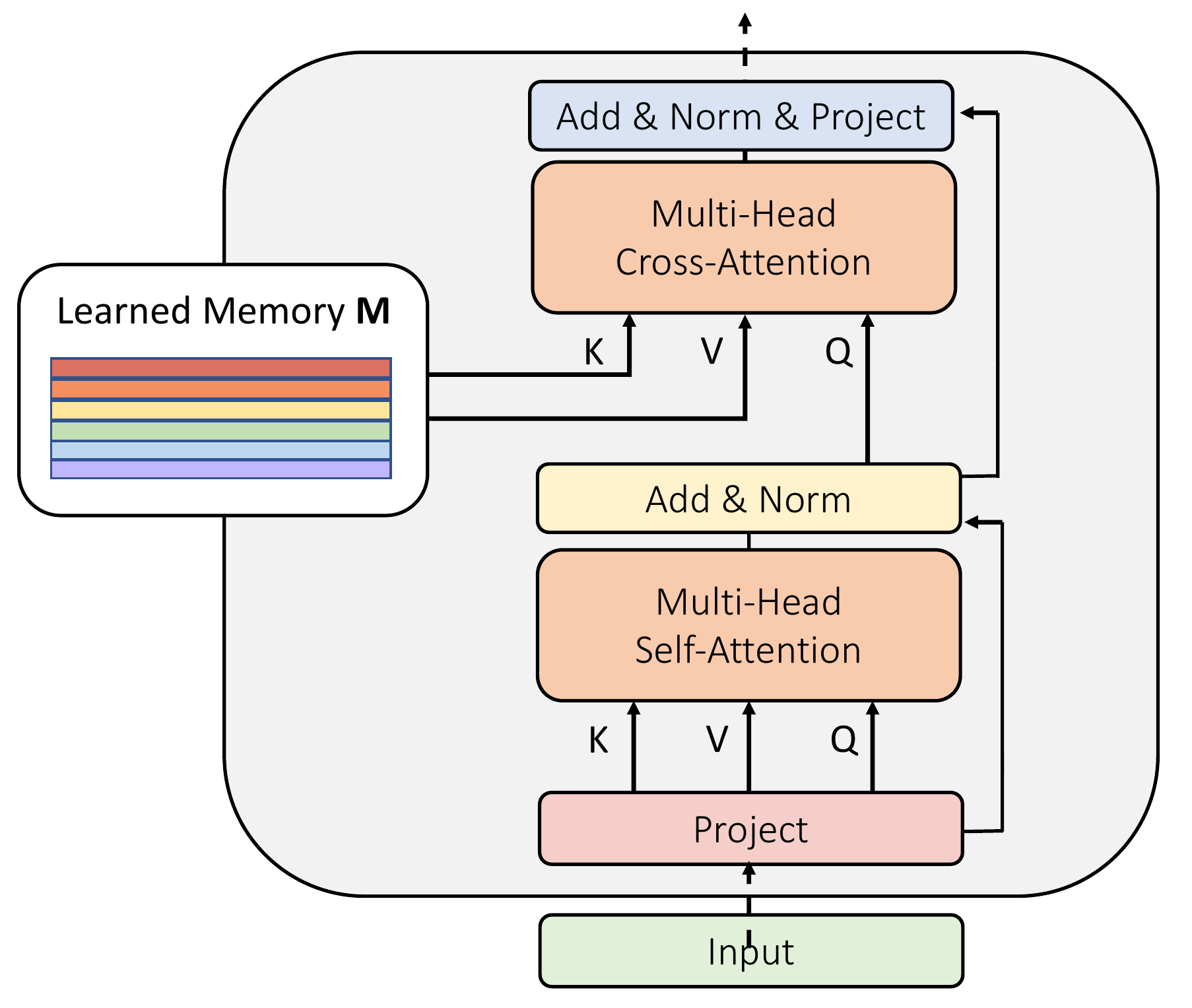}
\caption{Illustration of the transformer block with self-attention and cross-attention with memory. The learned memory, \textbf{M}, serves as an external memory bank and is globally shared across all transformer blocks. \text{K}, \text{V} and Q symbolize key, value and query, resp.
}\label{fig:attn}
\end{figure}

\vspace{5pt}
\noindent\textbf{Memory-augmented transformer decoder.} %
Our architecture is a stack of standard transformer decoder blocks~\cite{vaswani2017attention} (see Fig.~\ref{fig:attn}), where each such block has access to the global learnable memory, \eqref{eq:mem}.
Specifically, the memory-augmented transformer block consists of two key operations. First, the input is processed with the self-attention operation. Second, the cross-attention module attends to the learnable memory to generate the output. The input to self-attention in the first transformer block corresponds to the query, $\mathbf{Q}$. %
All cross-attention blocks access the same memory, $\textbf{M}$.
Intuitively, the memory module can be seen as a collection of learnable plan embeddings shared across the entire dataset. 
Empirically, we show that the memory module plays a key role in our framework.

Our transformer decoder, $\mathcal{T}$, consists of a stack of $N$ such memory-augmented blocks. We add two output heads (implemented as multi-layer perceptrons (MLPs)) at the final decoding layer. One head, $h_a$, yields intermediate predicted actions, $\tilde{a}_t$, while the second, $h_v$, yields corresponding intermediate visual representations, $\tilde{v}_t$, according to
\begin{equation}\label{eq:out}
  \tilde{a}_{1:T} = h_a(\mathcal{T}(\mathbf{Q},\mathbf{M})), \quad
  \tilde{v}_{1:T} = h_v(\mathcal{T}(\mathbf{Q}, \mathbf{M})),
\end{equation}

\subsection{Adversarial generative  modeling}\label{sec:gan}
To capture the uncertainty in prediction, where multiple plans from $v_{start}$ to $v_{goal}$ are plausible, we augment our model with a stochastic component using generative adversarial learning \cite{goodfellow2014generative}. To make the generation process conditional on the input, we augment the entire query input,~\eqref{eq:input}, with a random noise vector, $\mathit{z} \sim \mathcal{N}(0, 1), \mathit{z} \in \mathbb{R}^{d^\prime}$, through concatenation. The new query input sequence to our transformer, $\mathcal{T}$, thus becomes
\begin{equation}
\mathbf{Q}^z = \{[q_t; \mathit{z}] \ | \ q_t \in \mathbf{Q}\}.
\label{eq:input_noise}
\end{equation}

We employ adversarial training wherein the generator, $G$, is trained to produce realistic action sequences, while the critic, $C$, provides the supervisory signal for training $G$ \cite{arjovsky2017wasserstein}.
In our case, we treat the memory-augmented transformer, $\mathcal{T}$,  %
as the generator, $G$ (\ie, $G = h_v(\mathcal{T}(\mathbf{Q}^z, \mathbf{M}))$), while the critic is modeled by a simple MLP. More precisely, we pass the output of our transformer, $\tilde{v}_{1:T}$, concatenated along the temporal dimension, to the critic, $C$, which outputs a value between $0$ and $1$, indicating its ability to discriminate between the predicted and ground truth sequence, as depicted in Fig.~\ref{fig:alg}. Notably, to avoid the notorious issue of mode collapse associated with training GANs \cite{veegan} (\ie, regardless of variations in random latent noise, $\mathit{z}$), we follow previous work \cite{ndr1, ndr2, zhao2020diverse} and include the normalized distance regularizing loss, $\mathcal{L}_{reg}$, defined in the supplement.

\subsection{Training}\label{sec:losses}
To supervise our transformer, we rely on two complementary loss functions that enforce our transformer to decode the correct set of action labels in the procedure as well as corresponding visual representations. %
We also use an adversarial loss to train the stochastic component of our model.

\vspace{5pt}
\noindent\textbf{Visual step supervision.}
One of the outputs of our model at training time is the sequence of visual features, corresponding to %
the procedure steps, $\tilde{v}_{1:T}$. To supervise the visual features with corresponding language features, $l_{1:T}$, we adopt contrastive learning~\cite{GutmannH10}. For each feature, $\tilde{v}_t$, predicted by the transformer's head, $h_v$, we use the corresponding ground truth language embedding, $l_t$, as the positive example and all the other embeddings in the language vocabulary, $\{l_j\}$, as negative examples. The contrastive loss is calculated as
\begin{equation}
    \scalebox{1.05}{$\mathcal{L}_{l} = -\; \sum_{t=1}^{T} \left[ \text{log} \; \frac{\text{exp}(l_{t} \cdot \tilde{v}_{t} )}{\sum_{j} \text{exp}( l_{j} \cdot \tilde{v}_{t})} \right] $},
\end{equation}

\noindent where $(\cdot)$ denotes the dot-product operator.
Note, we use all examples in the language vocabulary as negatives as our vocabulary is typically small ($<1$K elements) and doing so allows for better training compared to per-batch negative sampling, \eg, \cite{HanXZ19}.\\

\noindent\textbf{Action plan supervision.}
We also enforce the action prediction head, $h_a$, to output sequences of action probabilities, $\tilde{a}_t$, corresponding to ground truth one-hot labels, $a_t$.
For this purpose, we use the cross-entropy loss
\begin{equation}
\scalebox{1.05}{$\mathcal{L}_{a} = -\sum_{t=1}^{T} \, a_t \, \text{log} \, {\tilde{a}_t}$.}
\end{equation}

\noindent\textbf{Adversarial supervision.} To model uncertainty, we use adversarial training on the visual state predictions, $\tilde{v}_{1:T}$. %
The goal is to make predicted visual observation sequences indistinguishable from feature sequences composed of the ground truth language step description, $l_{1:T}$.
We optimize the generator, $G$, (our transformer) and the critic, $C$, (an MLP) using an adversarial loss~\cite{arjovsky2017wasserstein}
\begin{equation}
\mathcal{L}_{adv} = \operatorname*{min}_{G} \operatorname*{max}_{C} \mathcal{V} (G, C, \mathbf{Q}^z, \mathbf{M}), 
\end{equation}
where $\mathcal{V}$ is the standard GAN objective, defined as $\mathbb{E}_{l \sim p_{data}}[\text{log}C(l)] + E_{z \sim p_z}[\text{log}(1-C(h_v(\mathcal{T}(\mathbf{Q}^z, \mathbf{M}))))]$, with $l {\small \sim} p_{data}$ and $z {\small \sim} p_z$ denoting the data distributions of the language representation and random noise, resp.\\

\noindent\textbf{Complete loss.} Overall, our full loss function is defined as
\begin{equation}
   \mathcal{L}(\theta) =  \lambda_1 \mathcal{L}_{l} + \lambda_2 \mathcal{L}_{\text{a}} + \lambda_3 \mathcal{L}_{adv} + \lambda_4 \mathcal{L}_{reg},
\end{equation}
where $\theta$ refers to the parameters associated with all learnable modules, \ie, queries, memory module, as well as the transformer decoder and discriminator parameters and $\lambda_{1:4}$ are empirically determined loss weights.

\subsection{Inference}\label{sec:inference}
At inference time, we use our transformer as a generative model to sample multiple procedure plans, $\tilde{\pi}^k = \tilde{a}^k_{1:T}$, %
for the same input start and goal observations.
This operation is achieved by drawing $\mathcal{K}$ latent noise vectors, $\mathit{z}^k$, and forwarding them through our transformer, $\mathcal{T}$, conditioned on a single start-goal observation, %
as follows
\begin{equation}
    \tilde{\pi}^k = h_a(\mathcal{T}(\mathbf{Q}^{z^k}, \mathbf{M})), \ \mathit{z}^k \sim \mathcal{N}(0, 1),
\end{equation}
for  $k = 1, \dotsc, \mathcal{K}$.

To obtain a probability distribution over actions at each timestep, $t$, of the plan, given by our model, we calculate action frequencies as follows:
\begin{equation}\label{eq:action_probs}
    \bar{\Pi} = \bar{a}_{1:T} = \frac{1}{\mathcal{K}} \sum_{k=1}^{\mathcal{K}} [\tilde{a}^k_1, \ldots, \tilde{a}^k_{T} ].
\end{equation}
Given that $a^k_t$ are one-hot vectors, each $\bar{a}_t$ results in a marginal distribution over actions at a specific timestep, $t$.

Most standard benchmark metrics for procedure planning, such as Success Rate, Accuracy or Intersection over Union (IoU), require a single action sequence output for evaluation (see Section~\ref{sec:metrics}).
To compute the most probable action sequence,
induced by our action distribution, $\bar{\Pi}$, we use the \textit{Viterbi} algorithm~\cite{viterbi}, as commonly seen in sequential labeling work \cite{richard2018neuralnetwork, koller2017re, koller2016deep, richard2017weakly}.
More specifically, we use $\bar{\Pi}$ as the emission matrix in the \textit{Viterbi} formulation, while the transition matrix is estimated from action co-occurrence frequencies in the training set (details in the supplement).
Our Viterbi post-processing step can be viewed as biasing sample selections from $\{\tilde{\pi}^k\}_{k=1}^{\mathcal{K}}$, toward plans that are more likely under a first-order model of action transitions.
An alternative approach to select a likely action sequence is simply to select the mode from the set $\{\tilde{\pi}^k\}_{k=1}^{\mathcal{K}}$. We empirically demonstrate the superiority of the Viterbi approach, which proved especially useful for smaller datasets.

\subsection{Implementation details}\label{sec:imp_details}

Our planner operates on video and language features, pre-extracted by a model trained for joint video-text embedding~\cite{miech2020end} using the HowTo100M~\cite{miech19howto100m} dataset and self supervision. We use a memory-augmented transformer with two layers and eight heads, and optimize it for 200 epochs with ADAM~\cite{kingma2014adam} on a single V100 GPU. Additional training and architecture details are provided in the supplement.

\section{Experiments}
In this section, we evaluate the role of each module in our approach (Sec.~\ref{sec:abl}) and demonstrate its performance across three different datasets. We include evaluation on the largest labeled instructional video dataset, which has not been used previously for the task of procedure planning due to the need of strong supervision in previous work (Sec.~\ref{sec:eval-all}).
Finally, we provide prediction uncertainty evaluation in procedure planning for the first time, which sheds light on our approach and the task of planning itself (Sec.~\ref{sec:eval-prob}).

\subsection{Evaluation protocol}
\noindent\textbf{Datasets.} For evaluation, we use three different instructional video datasets, namely, CrossTask \cite{CrossTask}, the Narrated Instructional Videos datasets \cite{alayrac2016unsupervised} (NIV) and COIN~\cite{COIN}. %
CrossTask contains 2750 videos, depicting 18 different procedure and an average of 7.6 actions/video; the NIV dataset is much smaller with 150 videos, five procedures and 9.5 actions/video on average. COIN is the largest dataset in our evaluation. It contains 11827 videos, 778 procedures and 3.6 actions/video.
Depicted procedures vary widely, \eg, %
\textit{Make Taco Salad} and \textit{Change Car Tire}.
We follow previous work \cite{procedure2020} and adopt 70\%/30\% to create our train/test splits and we use 20\% of the training data for validation. We also follow the data pre-processing steps outlined in the original procedure planning paper \cite{procedure2020} to select $\{\text{start}, \text{goal}\}$ observations and curate the dataset into plans covering different time horizons. More details are in the supplement.

\vspace{5pt}
\noindent\textbf{Metrics.}\label{sec:metrics} Following previous work~\cite{procedure2020, bi2021procedure, sun2021plate}, we evaluate the performance using three increasingly strict metrics. 
(i) mean Intersection over Union (mIoU) treats the predicted and ground truth action sequences as sets, and measures the overlap between these sets. mIoU is agnostic to the order of actions and only indicates whether the model captures the correct set of steps needed to complete the plan.
(ii) mean Accuracy (mAcc) performs element-wise comparisons between the predicted and ground truth action sequences, thereby considering the order of the actions as well. (iii) Success Rate (SR) considers a plan successful only if it \textit{exactly} matches the ground truth. %

We also evaluate the stochastic nature of our model by measuring the following probabilistic metrics: (i) the Kullback–Leibler (KL) divergence between our predicted plan distributions and ground truth; (ii) how well the ground truth modes are covered by our results (Mode Recall); (iii)  how often our plans correspond to the ground truth mode (Mode Precision). To this end, for each $\{\text{start}, \text{goal}\}$ observation, we draw ($\mathcal{K}$ = 1500) samples from our generative model and explicitly approximate a distribution, as described in Sec.~\ref{sec:inference}. For completeness, we also evaluate using more standard probabilistic prediction metrics, including Negative Log Likelihood (NLL) and the cosine distance \cite{dai2017towards, mehrasa2019variational, zhao2020diverse}.

\begin{table}[t]
\centering %
\resizebox{0.8\columnwidth}{!}{
\begin{tabular}{c| l c c c}
\hline %

Datasets & Loss Objective & SR $\uparrow$ & mAcc $\uparrow$ & mIoU $\uparrow$ \\ %
\hline %
\multirow{4}*{CrossTask}
 & $L_{a}$ & 16.90 & 44.20 & 57.56  \\ %
 & $L_{a}$ + $L_{l}$ & 22.12 & 45.57 & 67.40  \\ %
 & $L_{a}$ + $L_{l}$ + $L_{adv}$ & \textbf{23.34} & \textbf{49.96} & \textbf{73.89}  \\ %
 \cline{2-5}
 & w/o \textit{Viterbi} & 22.66 & 45.95 & 67.52  \\ %
\hline
\multirow{4}*{COIN}
& $L_{a}$& 8.48 & 12.19 & 68.15  \\ %
 & $L_{a}$ + $L_{l}$ & 14.41 & 20.25 & 73.49  \\ %
 & $L_{a}$ + $L_{l}$ + $L_{adv}$ & \textbf{15.40} & \textbf{21.67} & \textbf{76.31}  \\ %
 \cline{2-5}
 & w/o \textit{Viterbi} & 14.18 & 21.01 & 75.62  \\ %
\hline %
\multirow{4}*{NIV}
& $L_{a}$&  17.81 & 42.35 & 69.42  \\ %
 & $L_{a}$ + $L_{l}$& 24.05 & 46.67 & 73.89  \\ %
 & $L_{a}$ + $L_{l}$ + $L_{adv}$ & \textbf{24.68} & \textbf{49.01} & \textbf{74.29}  \\ %
 \cline{2-5}
 & w/o Viterbi & 20.18 & 47.73 & 73.09  \\ %
\hline %
\end{tabular}
}
\caption{
Performance of our model trained with different loss functions on three datasets. The last row of each block represents the results of the total loss but without the \textit{Viterbi} algorithm.
}
\label{tab:abl_viterbi}
\end{table}

\begin{table}[t]
	\centering
	\resizebox{0.6\columnwidth}{!}{
	\begin{tabular}{c c c c}
		\toprule %
		Memory Size & \multicolumn{1}{c}{SR $\uparrow$} & \multicolumn{1}{c}{mAcc $\uparrow$} & \multicolumn{1}{c}{mIoU $\uparrow$} \\ %
		\midrule %
		0 & 7.49 & 22.76 & 31.33 \\ %
		64 & 16.30 & 43.62 & 55.66 \\ %
		128 & \textbf{23.34} & \textbf{49.96} & \textbf{73.89} \\ %
		256 & 20.81 & 44.61 & 59.70 \\ %
		\bottomrule %
	\end{tabular}}
	\caption{
	Ablation study on the impact of external memory sizes for prediction horizon, $T=3$, with CrossTask. All results are obtained using our transformer, with two layers and eight heads.
	}
	\label{tab:abl_config}
\end{table}
\vspace{5pt}
\noindent\textbf{Baselines.} We compare to all previous approaches to procedure planning from instructional videos~\cite{procedure2020, sun2021plate, bi2021procedure} as well as other fully supervised planning approaches~\cite{ehsani2018let, abu2019uncertainty, srinivas2018universal}.

\subsection{Ablation study}\label{sec:abl}
\noindent\textbf{Impact of different loss functions.} We evaluate the role of each loss component by gradually introducing each objective. The results in Table~\ref{tab:abl_viterbi} show the pivotal role of language-based supervision, as evidenced by increased performance across all metrics, 
and the complementarity of the three objectives.
Notably, improvements from the adversarial loss 
may
seem 
marginal
(\eg, $\sim1\%$ 
in SR), as the metrics only compare a single prediction to a single ground truth plan. We show its strict superiority to deterministic models for modelling distributions of procedures in Sec.~\ref{sec:eval-prob}.

\vspace{5pt}
\noindent\textbf{Impact of Viterbi post-processing.} Throughout the empirical results, we use the Viterbi algorithm on top of the predicted action probabilities, $\bar{\Pi}$ \eqref{eq:action_probs}, to produce the optimal plan at inference time.
Notably, the Viterbi post-processing is optional, and one may directly use the set $\tilde{\pi}^k$ to produce the final plan by simply selecting the mode across the set. %
Comparing the last two rows in each block of Table~\ref{tab:abl_viterbi} shows the added advantage of using Viterbi to model the optimal action order in procedure plans explicitly, for all datasets.
Notably, Viterbi post-processing is especially helpful on the NIV dataset.
We hypothesize that the data scarcity in NIV results in a weaker predictive model at training; therefore, explicitly modeling the optimal transition between the actions with Viterbi plays a more important role in this case. %
\vspace{5pt}
\begin{table*}[t]
\centering %
\resizebox{0.7\textwidth}{!}{
\begin{tabular}{l c c c c c c c}
\toprule %
& & \multicolumn{3}{c}{$T=3$} & \multicolumn{3}{c}{$T=4$} \\ %
\cmidrule(l){3-5}  \cmidrule(l){6-8}

Models & \textit{Supervision} & SR $\uparrow$ & mAcc $\uparrow$ & mIoU $\uparrow$ & SR $\uparrow$ & mAcc $\uparrow$ & mIoU $\uparrow$    \\ %
\midrule %
Random & - & $<$0.01 & 0.94 & 1.66 & $<$0.01 & 0.83 & 1.66\\ %
Retrieval-Based & - & 8.05 & 23.30 & 32.06 & 3.95 & 22.22 & 36.97 \\ %
WLTDO \cite{ehsani2018let} & - & 1.87 & 21.64 & 31.70 & 0.77 & 17.92 & 26.43 \\ %
UAAA \cite{abu2019uncertainty} & - & 2.15 & 20.21 & 30.87 & 0.98 & 19.86 & 27.09 \\ %
UPN \cite{srinivas2018universal} & V & 2.89 & 24.39 & 31.56 & 1.19 & 21.59 & 27.85\\ %

DDN \cite{procedure2020} & V & 12.18 & 31.29 & 47.48 & 5.97 & 27.10 & 48.46 \\ %
Ext-GAILw/o Aug. \cite{bi2021procedure} & V & 18.01 & 43.86 & 57.16 &  - & - &  - \\ %
Ext-GAIL \cite{bi2021procedure} & V & 21.27 & 49.46 & 61.70 &  \textbf{16.41} & 43.05 &  60.93 \\ %
Ours  & L &\textbf{23.34} & \textbf{49.96} &\textbf{73.89} & 13.40 & \textbf{44.16} & \textbf{70.01} \\ %
\bottomrule %
\end{tabular}
}
\caption{Evaluation of procedure planning results on CrossTask for prediction horizon $T \in \{3, 4\}$. The column name \textit{Supervision} denotes the type of state supervision applied in training, with V and L denoting visual and language state representation, resp.
}
\label{tab:crosstask_rst1}
\vspace{-6pt}
\end{table*}
\noindent\textbf{Impact of model configuration.} We also include an ablation evaluating the adopted memory augmented transformer decoder. 
Table~\ref{tab:abl_config} shows that the size of our memory plays a key role in our architecture. Indeed, excluding the memory component yields the worst results, while too large a memory degrades performance. These results suggest that the memory component helps capture dataset content, where it yields stronger results when the number of memory entries is large enough to properly span the actions present in the entire dataset. Notably, while tuning the memory size for each dataset might yield better results, for simplicity, we elect to use the best setting of CrossTask for all datasets.
\subsection{Comparison to alternative approaches}\label{sec:eval-all}
\begin{table}[t]
\centering %
\resizebox{0.8\columnwidth}{!}{
\begin{tabular}{l c c c c}
\toprule %
& \multicolumn{1}{c}{$T=3$} & \multicolumn{1}{c}{$T=4$} & \multicolumn{1}{c}{$T=5$} & \multicolumn{1}{c}{$T=6$}\\ %
\cmidrule(l){2-5}

Models & SR$\uparrow$ & SR$\uparrow$ & SR$\uparrow$ & SR$\uparrow$ \\ %
\midrule %
Retrieval-Based & 8.05 & 3.95 & 2.40 & 1.10 \\ %
DDN \cite{procedure2020} &  12.18 & 5.97 & 3.10 & 1.20 \\ %
Ours (Protocol 1) & \textbf{23.34} & \textbf{13.40} &\textbf{7.21} & \textbf{4.40} \\ %
\midrule
PlaTe \cite{sun2021plate} &  18.5 & 14.0 & 10.0 & 7.5 \\ %
Ours (Protocol 2) & \textbf{24.4} & \textbf{15.8} &\textbf{11.8} & \textbf{8.3} \\ %
\bottomrule %

\end{tabular}
}
\caption{
Success Rate evaluation of procedure planning results on CrossTask~\cite{CrossTask} that extends to longer prediction horizon, $T$. %
\label{tab:crosstask_rst2}
}
\end{table}
\noindent\textbf{CrossTask (short-horizon).} Table~\ref{tab:crosstask_rst1} compares our weakly-supervised approach to a number of alternatives, including the fully supervised state of the art, across the two prediction horizons typically reported in this task.
Our results are consistently better, except for the the success rate (SR) at $T=4$, where we are the second best approach. The performance improvement in SR is especially striking at short-term horizon, $T = 3$, where we outperform the previous best (\ie, Ext-GAIL~\cite{bi2021procedure}) by more than $2\%$, while using weaker supervision. Notably, Ext-GAIL achieves its level of performance via data augmentation, which allows it to have 30\% more training data.
In a more similar setup (\ie, when Ext-GAIL does not use data augmentation) the performance gain of our method over ``Ext-GAIL w/o Aug" is $5.3\%$. Importantly, our results are obtained with weaker supervision, which speaks decisively in favor of our approach.

We also notice a larger gain in mIoU compared to previous work, \ie, $73.89\% \; vs. \;61.70\%$ for $T =3$ and $70.01\% \; vs. \; 60.93\%$ for $T=4$. %
This result suggests that our approach is better at capturing feasible action steps than other approaches (\eg, never producing \textit{pour water} when input observations are related to \textit{making a salad}). We hypothesize that this performance is enabled by the language-based contrastive learning, which is %
more effective at clustering latent representations than its vision counterpart. For example, while some visual observations can look similar (\eg, \textit{pour water} and \textit{add oil}), the distinction between the two is clearer in natural language.
Notably, the level of improvements in mIoU and mAcc is typically higher than the gain in SR. We attribute this result to the uncertainty inherent to the task (\ie, multiple feasible plans for the same start and goal observations). We explore this aspect in greater detail in Sec.~\ref{sec:eval-prob}. Finally, thanks to the non-autoregressive nature, we are \textbf{4x} faster at inference, \eg, \textbf{6.75ms} (ours) vs. \textbf{27.34ms} (DDN~\cite{procedure2020}) on CrossTask for $T = 3$.

\begin{table}[t]
\centering %
\resizebox{\columnwidth}{!}{

\begin{tabular}{l l c c c c c c c}
\toprule %
& & & \multicolumn{3}{c}{NIV} & \multicolumn{3}{c}{COIN} \\ %
\cmidrule(l){4-6}  \cmidrule(l){7-9}
Horizons & Methods & Sup. & SR$\uparrow$ & mAcc$\uparrow$ & mIoU$\uparrow$ & SR$\uparrow$ & mAcc$\uparrow$ & mIoU$\uparrow$ \\ %
\midrule
\multirow{5}*{$T=3$}
 & Random & -& 2.21 & 4.07 &  6.09 & $<$0.01 & $<$0.01 & 2.47  \\ %
 & Retrieval & - & - & - &  - & 4.38 & 17.40 & 32.06  \\ %
  & DDN~\cite{procedure2020} & V & 18.41 & 32.54 & 56.56  & 13.9 & 20.19 & 64.78 \\ %
  & Ext-GAIL~\cite{bi2021procedure} & V & 22.11 & 42.20 & 65.93 & - & - & -\\ %
  & Ours & L & \textbf{24.68} & \textbf{49.01} & \textbf{74.29} & \textbf{15.4} & \textbf{21.67} & \textbf{76.31} \\ %
\midrule
\multirow{5}*{$T=4$}
 & Random & - & 1.12 & 2.73 & 5.84  &$<$0.01& $<$0.01&2.32\\ %
 & Retrieval & - & - & - & - & 2.71& 14.29& 36.97\\ %
  & DDN~\cite{procedure2020} & V & 15.97 & 27.09 & 53.84 & 11.13 & 17.71 & 68.06 \\ %
  & Ext-GAIL~\cite{bi2021procedure} & V &  19.91 & 36.31 & 53.84 & - & -&- \\ %
  & Ours & L & \textbf{20.14} & \textbf{38.36} & \textbf{67.29} & \textbf{11.32}& \textbf{18.85} & \textbf{70.53}\\ %
\bottomrule %
\end{tabular}
}
\caption{Procedure planning results on NIV~\cite{alayrac2016unsupervised} and COIN~\cite{COIN} for prediction horizon $T \in \{3, 4\}$. The column \textit{Sup.}~denotes the type of state supervision applied in training. 
}
\label{tab:coin_rst1}
\end{table}
\vspace{3pt}
\noindent\textbf{CrossTask (long-horizon).}
We now evaluate our model's ability to predict plans for longer time horizons (\ie, $T \in \{3, \dotsc, 6\}$). 
We compare to previous approaches that reported results on such horizons.  %
There are two different protocols for these settings. (i) Protocol~1~\cite{procedure2020} and (ii) Protocol~2~\cite{sun2021plate}; see details in supplement. %
For a fair comparison, we present our results using both protocols in Table~\ref{tab:crosstask_rst2} and show that our approach is the most effective on both. %

\vspace{3pt}
\noindent\textbf{NIV.} Following previous work~\cite{bi2021procedure}, we also evaluate our model on the smaller NIV dataset. The results in Table~\ref{tab:coin_rst1} are consistent with our results on CrossTask. Once again, our approach is the top performer across all metrics.
This result suggests our language supervision works for smaller datasets as well. %

\vspace{3pt}
\noindent\textbf{COIN.} To show the capability of our method to scale, we evaluate our method on the largest labeled instructional video dataset (\ie, COIN). %
As we are the first ones to perform procedure planning on such a large-scale dataset, there is a lack of comparison methods. Therefore, we follow existing work~\cite{procedure2020} and include three baselines: (i) Random selection; (ii) Retrieval-based; (iii) and a re-implementation of the DDN model~\cite{procedure2020} %
, using our video features, described in Sec.~\ref{sec:imp_details}. 
Table~\ref{tab:coin_rst1} shows that our model equipped with language supervision consistently outperforms the baselines, even strongly-supervised ones. %

\subsection{Evaluating probabilistic modeling}\label{sec:eval-prob}
To evaluate our probabilistic modeling, we compare plan distributions produced by our model, with the ground truth distribution over feasible plans.
We focus our evaluation of probabilistic modeling on CrossTask as it is the most suitable dataset in terms of variations in the set of feasible plans, as we demonstrate in the supplement.

\vspace{5pt}
\noindent\textbf{Plan distribution modeling.} Our approach is probabilistic by design as described in Secs.~\ref{sec:gan} and \ref{sec:inference}. %
To establish a deterministic baseline, we train our model without the adversarial loss and fix the latent noise vector, $\mathit{z}=0$, both during training and testing.
To build the ground truth distribution over goal-conditioned plans, we retrieve all action sequences of length $T$ in the test set that share the given start and goal state. %
The plan distribution of our probabilistic model (conditioned on the start and goal observations) is obtained by sampling $\mathcal{K}=1500$ different action sequences, as discussed in Sec.~\ref{sec:inference}.
For the deterministic baseline, the model produces only one plan (\ie, $\mathcal{K}=1$).
In all cases, the probability of a plan is defined as its frequency in the obtained sample set.
To evaluate the quality of the predicted plans, we measure the (dis-)similarity between plan distributions produced by each model (\ie, ours vs.~the deterministic variant) and that of the ground truth using the KL divergence and NLL.
Table~\ref{tab:prob_eval} shows that 
our probabilistic approach better matches the ground truth plan distribution (\ie, it has lower KL and NLL).
These results come about because our model is able to sample multiple \emph{valid} plans  with respect to the test set distribution, as opposed to the deterministic model that considers a single feasible plan.

\begin{table}[t]
\centering
\resizebox{\columnwidth}{!}{  
\begin{tabular}{l c c c c c}
\toprule %
Metric $\downarrow$ & Method & $T=3$ & $T=4$ & $T=5$ & $T=6$ \\ %
\midrule %
\multirow{2}*{KL-Div}
& Ours - determinstic & 2.31 & 4.47 & 6.30 & 8.81 \\ %
& Ours - probabilistic & \textbf{2.11} & \textbf{3.50} & \textbf{4.26} & \textbf{6.89} \\ %
\midrule %
\multirow{2}*{NLL}
& Ours - determinstic & 5.13 & 6.25 & 6.68 & 8.49 \\ %
& Ours - probabilistic & \textbf{4.89} & \textbf{5.48} & \textbf{6.24} & \textbf{7.67} \\ %
\bottomrule %
\end{tabular}
}
\caption{Evaluation of the plan distributions produced by our probabilistic approach vs. the deterministic variant.}
\label{tab:prob_eval}
\end{table}

\begin{figure}[t]
\centering
\scalebox{1}{ %
\includegraphics[scale=0.25]{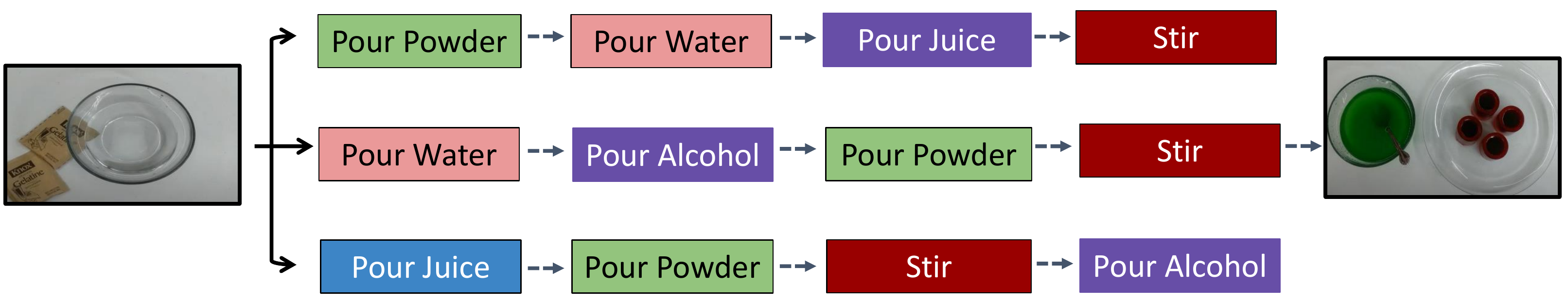}}
\caption{Sample \emph{plausible} plans (seen in the test set) produced by our probabilistic model, for the same \{start, goal\} observations.}%
\label{fig:diverse}
\vspace{-3pt}
\end{figure}

\vspace{5pt}
\noindent\textbf{Sample diversity and mode coverage.}
For given start and goal observations, our approach produces multiple plan hypotheses, using probabilistic sampling described in Sec.~\ref{sec:inference} (as visualized in Figure~\ref{fig:diverse}).
In this section, we measure the diversity of our samples and their relation to the ground truth plans distribution. To characterize the ground truth distribution, we define ground truth modes as the set of unique action sequences in the test set that share the same start and goal state. To calculate the relation of our samples to the ground truth modes we define two metrics, Mode Recall (ModeRec) and Mode Precision (ModePrec).
ModeRec reflects how well the GT modes are covered by our model and is calculated as the average number of GT modes captured by at least one sample from our model.
In complement, ModePrec measures how often a sampled plan is feasible according to the test data; it is calculated as the average number of samples that match at least one GT mode. 
Intuitively, ModeRec and ModeRec measure not only how diverse the samples from our model are, but also how useful this diversity is with respect to the GT. %
By measuring the average cosine distance among model samples, \ie, CosDist --- a metric widely used in the GAN literature --- we also show that our samples are diverse agnostic to the data distribution. 
The results in Table~\ref{tab:mode_eval} suggest our probabilistic approach can produce both diverse and accurate plans, where it is superior on all metrics.

\begin{table}[t]
\centering
\resizebox{\columnwidth}{!}{
\begin{tabular}{l l c c c c}
\toprule %
Metric $\uparrow$ & Method  & \multicolumn{1}{c}{$T=3$} & \multicolumn{1}{c}{$T=4$} & \multicolumn{1}{c}{$T=5$} & \multicolumn{1}{c}{$T=6$}\\ %
\midrule
\multirow{2}*{ModePrec}
& Ours - deterministic & 27.61 & 17.21 & 7.41  & 4.97 \\ %
& Ours - probabilistic & \textbf{36.61} & \textbf{18.55} & \textbf{12.48}  & \textbf{6.58} \\ %
\midrule
\multirow{2}*{ModeRec}
& Ours - deterministic  & 56.24 & 37.33 & 18.38  & 8.85 \\ %
& Ours - probabilistic  & \textbf{66.13} & \textbf{46.56} & \textbf{26.46}  & \textbf{12.67} \\ %
\midrule
CosDist & Ours - probabilistic & 0.384 & 0.302 & 0.2471 & 0.1658 \\ %

\bottomrule %
\end{tabular}}
\caption{Evaluation of diversity and accuracy of our samples with respect to ground truth. Our approach improves \emph{both} ModePrec and ModeRec metrics. We further provide averaged pair-wise cosine distance as another indicator for diversity.
CosDist for the deterministic model is not provided, since it only produces a single result with no pairs to compare.
}%
\label{tab:mode_eval}
\end{table}

\vspace{-3pt}
\section{Conclusion}
We have introduced a weakly-supervised method for probabilistic procedure planning using instructional videos.
Different from previous work, we echew the need for expensive visual supervision in favor of cheaper language supervision by capitalizing on pre-trained text-video embeddings, which, remarkably, leads to superior planning performance.
We showed that modeling the interplay between intermediate visual states and actions step-by-step is not a necessity %
for procedure planning. Instead, we efficiently solve the problem with a ``one-shot'' transformer decoder architecture. In addition, we demonstrated the crucial role of modeling uncertainty in obtained plans to yield a principled approach to planning from videos.
We introduced a way to evaluate such uncertainty on the test set and show that it is a powerful metric to better understand the model and the task of planning itself.
Hopefully, future work will adopt the probabilistic view on procedure planning from instructional videos, not only in training but also in evaluation, such that the next generation of planners can confidently predict multiple feasible plans to achieve the desired goal.

\newpage

{\small
\bibliographystyle{ieee_fullname}
\bibliography{bibref_long,bibref}

\begin{thebibliography}{10}\itemsep=-1pt

\bibitem{abu2019uncertainty}
Yazan Abu~Farha and Juergen Gall.
\newblock Uncertainty-aware anticipation of activities.
\newblock In {\em Proceedings of the International Conference on Computer
  Vision (ICCV)}, 2019.

\bibitem{alayrac2016unsupervised}
Jean-Baptiste Alayrac, Piotr Bojanowski, Nishant Agrawal, Josef Sivic, Ivan
  Laptev, and Simon Lacoste-Julien.
\newblock Unsupervised learning from narrated instruction videos.
\newblock In {\em Proceedings of the IEEE Conference on Computer Vision and
  Pattern Recognition (CVPR)}, 2016.

\bibitem{arjovsky2017wasserstein}
Martin Arjovsky, Soumith Chintala, and L{\'e}on Bottou.
\newblock Wasserstein generative adversarial networks.
\newblock In {\em International Conference on Machine Learning (ICML)}, 2017.

\bibitem{arnab2021vivit}
Anurag Arnab, Mostafa Dehghani, Georg Heigold, Chen Sun, Mario Lu{\v{c}}i{\'c},
  and Cordelia Schmid.
\newblock {ViVit}: A video vision transformer.
\newblock {\em arXiv preprint arXiv:2103.15691}, 2021.

\bibitem{bello2019attention}
Irwan Bello, Barret Zoph, Ashish Vaswani, Jonathon Shlens, and Quoc~V Le.
\newblock Attention augmented convolutional networks.
\newblock In {\em Proceedings of the International Conference on Computer
  Vision (ICCV)}, 2019.

\bibitem{bi2021procedure}
Jing Bi, Jiebo Luo, and Chenliang Xu.
\newblock Procedure planning in instructional videos via contextual modeling
  and model-based policy learning.
\newblock In {\em Proceedings of the International Conference on Computer
  Vision (ICCV)}, 2021.

\bibitem{carion2020end}
Nicolas Carion, Francisco Massa, Gabriel Synnaeve, Nicolas Usunier, Alexander
  Kirillov, and Sergey Zagoruyko.
\newblock End-to-end object detection with transformers.
\newblock In {\em Proceedings of the European Conference on Computer Vision
  (ECCV)}, 2020.

\bibitem{procedure2020}
Chien-Yi Chang, De-An Huang, Danfei Xu, Ehsan Adeli, Li Fei-Fei, and
  Juan~Carlos Niebles.
\newblock Procedure planning in instructional videos.
\newblock In {\em Proceedings of the European Conference on Computer Vision
  (ECCV)}, 2020.

\bibitem{dai2017towards}
Bo Dai, Sanja Fidler, Raquel Urtasun, and Dahua Lin.
\newblock Towards diverse and natural image descriptions via a conditional
  {GAN}.
\newblock In {\em Proceedings of the International Conference on Computer
  Vision (ICCV)}, 2017.

\bibitem{desai2021virtex}
Karan Desai and Justin Johnson.
\newblock {VirTex}: Learning visual representations from textual annotations.
\newblock In {\em Proceedings of the IEEE Conference on Computer Vision and
  Pattern Recognition (CVPR)}, 2021.

\bibitem{drop-dtw}
Nikita Dvornik, Isma Hadji, Konstantinos Derpanis, Animesh Garg, and Allan
  Jepson.
\newblock {Drop-DTW}: Aligning common signal between sequences while dropping
  outliers.
\newblock In {\em Advances in Neural Information Processing Systems (NeurIPS)},
  2021.

\bibitem{ehsani2018let}
Kiana Ehsani, Hessam Bagherinezhad, Joseph Redmon, Roozbeh Mottaghi, and Ali
  Farhadi.
\newblock Who let the dogs out? {M}odeling dog behavior from visual data.
\newblock In {\em Proceedings of the IEEE Conference on Computer Vision and
  Pattern Recognition (CVPR)}, 2018.

\bibitem{finn2017deep}
Chelsea Finn and Sergey Levine.
\newblock Deep visual foresight for planning robot motion.
\newblock In {\em Proceedings of the IEEE International Conference on Robotics
  and Automation (ICRA)}, 2017.

\bibitem{florensa2018automatic}
Carlos Florensa, David Held, Xinyang Geng, and Pieter Abbeel.
\newblock Automatic goal generation for reinforcement learning agents.
\newblock In {\em International Conference on Machine Learning (ICML)}, 2018.

\bibitem{gabeur2020multi}
Valentin Gabeur, Chen Sun, Karteek Alahari, and Cordelia Schmid.
\newblock Multi-modal transformer for video retrieval.
\newblock In {\em Proceedings of the European Conference on Computer Vision
  (ECCV)}, 2020.

\bibitem{ghosh2018learning}
Dibya Ghosh, Abhishek Gupta, and Sergey Levine.
\newblock Learning actionable representations with goal-conditioned policies.
\newblock {\em arXiv preprint arXiv:1811.07819}, 2018.

\bibitem{gomez2017self}
Lluis Gomez, Yash Patel, Mar{\c{c}}al Rusi{\~n}ol, Dimosthenis Karatzas, and CV
  Jawahar.
\newblock Self-supervised learning of visual features through embedding images
  into text topic spaces.
\newblock In {\em Proceedings of the IEEE Conference on Computer Vision and
  Pattern Recognition (CVPR)}, 2017.

\bibitem{goodfellow2014generative}
Ian Goodfellow, Jean Pouget-Abadie, Mehdi Mirza, Bing Xu, David Warde-Farley,
  Sherjil Ozair, Aaron Courville, and Yoshua Bengio.
\newblock Generative adversarial nets.
\newblock In {\em Advances in Neural Information Processing Systems (NeurIPS)},
  2014.

\bibitem{GutmannH10}
Urs~Michael Gutmann and Aapo Hyv{\"a}rinen.
\newblock Noise-contrastive estimation: {A} new estimation principle for
  unnormalized statistical models.
\newblock In {\em Proceedings of the International Conference on Artificial
  Intelligence and Statistics (AISTATS)}, 2010.

\bibitem{HanXZ19}
Tengda Han, Weidi Xie, and Andrew Zisserman.
\newblock Video representation learning by dense predictive coding.
\newblock In {\em Proceedings of the International Conference on Computer
  Vision Workshops (ICCVW)}, 2019.

\bibitem{wikihow}
Jack Herrick.
\newblock {wikiHow}.
\newblock \url{www.wikihow.com}, 2005.
\newblock [Online how-to website].

\bibitem{joulin2016learning}
Armand Joulin, Laurens Van Der~Maaten, Allan Jabri, and Nicolas Vasilache.
\newblock Learning visual features from large weakly supervised data.
\newblock In {\em Proceedings of the European Conference on Computer Vision
  (ECCV)}, 2016.

\bibitem{kaelbling1993hierarchical}
Leslie~Pack Kaelbling.
\newblock Hierarchical learning in stochastic domains: Preliminary results.
\newblock In {\em International Conference on Machine Learning (ICML)}, 1993.

\bibitem{kingma2014adam}
Diederik~P Kingma and Jimmy Ba.
\newblock Adam: A method for stochastic optimization.
\newblock {\em arXiv preprint arXiv:1412.6980}, 2014.

\bibitem{vae}
Diederik~P Kingma and Max Welling.
\newblock Auto-encoding variational bayes.
\newblock In {\em International Conference on Machine Learning (ICML)}, 2014.

\bibitem{dosovitskiy2020image}
Alexander Kolesnikov, Alexey Dosovitskiy, Dirk Weissenborn, Georg Heigold,
  Jakob Uszkoreit, Lucas Beyer, Matthias Minderer, Mostafa Dehghani, Neil
  Houlsby, Sylvain Gelly, Thomas Unterthiner, and Xiaohua Zhai.
\newblock An image is worth 16x16 words: Transformers for image recognition at
  scale.
\newblock In {\em International Conference on Learning Representations (ICLR)},
  2021.

\bibitem{koller2016deep}
Oscar Koller, Hermann Ney, and Richard Bowden.
\newblock Deep hand: How to train a {CNN} on 1 million hand images when your
  data is continuous and weakly labelled.
\newblock In {\em Proceedings of the IEEE Conference on Computer Vision and
  Pattern Recognition (CVPR)}, 2016.

\bibitem{koller2017re}
Oscar Koller, Sepehr Zargaran, and Hermann Ney.
\newblock Re-sign: Re-aligned end-to-end sequence modelling with deep recurrent
  {CNN-HMMs}.
\newblock In {\em Proceedings of the IEEE Conference on Computer Vision and
  Pattern Recognition (CVPR)}, 2017.

\bibitem{lei2020mart}
Jie Lei, Liwei Wang, Yelong Shen, Dong Yu, Tamara Berg, and Mohit Bansal.
\newblock {MART}: Memory-augmented recurrent transformer for coherent video
  paragraph captioning.
\newblock In {\em Proceedings of the 58th Annual Meeting of the Association for
  Computational Linguistics}, pages 2603--2614, 2020.

\bibitem{ndr1}
Shaohui Liu, Xiao Zhang, Jianqiao Wangni, and Jianbo Shi.
\newblock Normalized diversification.
\newblock In {\em Proceedings of the IEEE Conference on Computer Vision and
  Pattern Recognition (CVPR)}, 2019.

\bibitem{vilbert}
Jiasen Lu, Dhruv Batra, Devi Parikh, and Stefan Lee.
\newblock {ViLBERT}: Pretraining task-agnostic visiolinguistic representations
  for vision-and-language tasks.
\newblock In {\em Advances in Neural Information Processing Systems (NeurIPS)},
  2019.

\bibitem{Luo2020UniVL}
Huaishao Luo, Lei Ji, Botian Shi, Haoyang Huang, Nan Duan, Tianrui Li, Jason
  Li, Taroon Bharti, and Ming Zhou.
\newblock {U}ni{VL}: A unified video and language pre-training model for
  multimodal understanding and generation.
\newblock {\em arXiv preprint arXiv:2002.06353}, 2020.

\bibitem{mangalam2020not}
Karttikeya Mangalam, Harshayu Girase, Shreyas Agarwal, Kuan-Hui Lee, Ehsan
  Adeli, Jitendra Malik, and Adrien Gaidon.
\newblock It is not the journey but the destination: Endpoint conditioned
  trajectory prediction.
\newblock In {\em Proceedings of the European Conference on Computer Vision
  (ECCV)}, 2020.

\bibitem{marchetti2020mantra}
Francesco Marchetti, Federico Becattini, Lorenzo Seidenari, and Alberto~Del
  Bimbo.
\newblock {MANTRA}: Memory augmented networks for multiple trajectory
  prediction.
\newblock In {\em Proceedings of the IEEE Conference on Computer Vision and
  Pattern Recognition (CVPR)}, 2020.

\bibitem{mehrasa2019variational}
Nazanin Mehrasa, Akash~Abdu Jyothi, Thibaut Durand, Jiawei He, Leonid Sigal,
  and Greg Mori.
\newblock A variational auto-encoder model for stochastic point processes.
\newblock In {\em Proceedings of the IEEE Conference on Computer Vision and
  Pattern Recognition (CVPR)}, 2019.

\bibitem{miech2020end}
Antoine Miech, Jean-Baptiste Alayrac, Lucas Smaira, Ivan Laptev, Josef Sivic,
  and Andrew Zisserman.
\newblock End-to-end learning of visual representations from uncurated
  instructional videos.
\newblock In {\em Proceedings of the IEEE Conference on Computer Vision and
  Pattern Recognition (CVPR)}, 2020.

\bibitem{miech19howto100m}
Antoine Miech, Dimitri Zhukov, Jean-Baptiste Alayrac, Makarand Tapaswi, Ivan
  Laptev, and Josef Sivic.
\newblock How{T}o100{M}: Learning a text-video embedding by watching hundred
  million narrated video clips.
\newblock In {\em Proceedings of the International Conference on Computer
  Vision (ICCV)}, 2019.

\bibitem{pashevich2021episodic}
Alexander Pashevich, Cordelia Schmid, and Chen Sun.
\newblock Episodic transformer for vision-and-language navigation.
\newblock In {\em Proceedings of the International Conference on Computer
  Vision (ICCV)}, 2021.

\bibitem{piergiovanni2020adversarial}
AJ Piergiovanni, Anelia Angelova, Alexander Toshev, and Michael~S Ryoo.
\newblock Adversarial generative grammars for human activity prediction.
\newblock In {\em Proceedings of the European Conference on Computer Vision
  (ECCV)}, 2020.

\bibitem{radford2021learning}
Alec Radford, Jong~Wook Kim, Chris Hallacy, Aditya Ramesh, Gabriel Goh,
  Sandhini Agarwal, Girish Sastry, Amanda Askell, Pamela Mishkin, Jack Clark,
  Krueger Gretchen, and Sutskever Ilya.
\newblock Learning transferable visual models from natural language
  supervision.
\newblock In {\em International Conference on Machine Learning (ICML)}, 2021.

\bibitem{richard2017weakly}
Alexander Richard, Hilde Kuehne, and Juergen Gall.
\newblock Weakly supervised action learning with {RNN} based fine-to-coarse
  modeling.
\newblock In {\em Proceedings of the IEEE Conference on Computer Vision and
  Pattern Recognition (CVPR)}, pages 754--763, 2017.

\bibitem{richard2018neuralnetwork}
Alexander Richard, Hilde Kuehne, Ahsan Iqbal, and Juergen Gall.
\newblock {NeuralNetwork-Viterbi}: A framework for weakly supervised video
  learning.
\newblock In {\em Proceedings of the IEEE Conference on Computer Vision and
  Pattern Recognition (CVPR)}, 2018.

\bibitem{rnn}
David~E Rumelhart, Geoffrey~E Hinton, and Ronald~J Williams.
\newblock Learning representations by back-propagating errors.
\newblock {\em Nature}, 323(6088):533--536, 1986.

\bibitem{srinivas2018universal}
Aravind Srinivas, Allan Jabri, Pieter Abbeel, Sergey Levine, and Chelsea Finn.
\newblock Universal planning networks: Learning generalizable representations
  for visuomotor control.
\newblock In {\em International Conference on Machine Learning (ICML)}, 2018.

\bibitem{veegan}
Akash Srivastava, Lazar Valkov, Chris Russell, Michael~U Gutmann, and Charles
  Sutton.
\newblock {VEEGAN}: Reducing mode collapse in gans using implicit variational
  learning.
\newblock In {\em Advances in Neural Information Processing Systems (NeurIPS)},
  2017.

\bibitem{sun2021plate}
Jiankai Sun, De-An Huang, Bo Lu, Yun-Hui Liu, Bolei Zhou, and Animesh Garg.
\newblock {PlaTe}: Visually-grounded planning with transformers in procedural
  tasks.
\newblock {\em arXiv preprint arXiv:2109.04869v1}, 2021.

\bibitem{COIN}
Yansong Tang, Dajun Ding, Yongming Rao, Yu Zheng, Danyang Zhang, Lili Zhao,
  Jiwen Lu, and Jie Zhou.
\newblock {COIN}: A large-scale dataset for comprehensive instructional video
  analysis.
\newblock In {\em Proceedings of the IEEE Conference on Computer Vision and
  Pattern Recognition (CVPR)}, 2019.

\bibitem{vaswani2017attention}
Ashish Vaswani, Noam Shazeer, Niki Parmar, Jakob Uszkoreit, Llion Jones,
  Aidan~N Gomez, {\L}ukasz Kaiser, and Illia Polosukhin.
\newblock Attention is all you need.
\newblock In {\em Advances in Neural Information Processing Systems (NeurIPS)},
  2017.

\bibitem{viterbi}
Andrew Viterbi.
\newblock Error bounds for convolutional codes and an asymptotically optimum
  decoding algorithm.
\newblock {\em IEEE Transactions on Information Theory}, 13(2):260--269, 1967.

\bibitem{wu2020memformer}
Qingyang Wu, Zhenzhong Lan, Jing Gu, and Zhou Yu.
\newblock Memformer: The memory-augmented transformer.
\newblock {\em arXiv preprint arXiv:2010.06891}, 2020.

\bibitem{videoclip}
Hu Xu, Gargi Ghosh, Po-Yao Huang, Dmytro Okhonko, Armen Aghajanyan, Florian
  Metze, Luke Zettlemoyer, and Christoph Feichtenhofer.
\newblock {VideoCLIP}: Contrastive pre-training for zero-shot video-text
  understanding.
\newblock In {\em Proceedings of the Conference on Empirical Methods in Natural
  Language Processing (EMNLP)}, 2021.

\bibitem{ndr2}
Dingdong Yang, Seunghoon Hong, Yunseok Jang, Tianchen Zhao, and Honglak Lee.
\newblock Diversity-sensitive conditional generative adversarial networks.
\newblock In {\em International Conference on Learning Representations (ICLR)},
  2018.

\bibitem{yuan2020dlow}
Ye Yuan and Kris Kitani.
\newblock {DLow}: Diversifying latent flows for diverse human motion
  prediction.
\newblock In {\em Proceedings of the European Conference on Computer Vision
  (ECCV)}, 2020.

\bibitem{zhang2021temporal}
Chuhan Zhang, Ankush Gupta, and Andrew Zisserman.
\newblock Temporal query networks for fine-grained video understanding.
\newblock In {\em Proceedings of the IEEE Conference on Computer Vision and
  Pattern Recognition (CVPR)}, 2021.

\bibitem{zhao2020diverse}
He Zhao and Richard~P Wildes.
\newblock On diverse asynchronous activity anticipation.
\newblock In {\em Proceedings of the European Conference on Computer Vision
  (ECCV)}, 2020.

\bibitem{CrossTask}
Dimitri Zhukov, Jean-Baptiste Alayrac, Ramazan~Gokberk Cinbis, David Fouhey,
  Ivan Laptev, and Josef Sivic.
\newblock Cross-task weakly supervised learning from instructional videos.
\newblock In {\em Proceedings of the IEEE Conference on Computer Vision and
  Pattern Recognition (CVPR)}, 2019.

\end{thebibliography}
}

\clearpage
\appendix
\section{Supplementary material}\label{sec:summary}
Our supplemental material is organized as follows: Section~\ref{sec:supp-approach} provides details on the regularization loss used in our adversarial training. Section~\ref{sec:sup-viterbi} elaborates on our inference procedure using the Viterbi algorithm. Section~\ref{sec:impl} includes a thorough description of our implementation details. We then provide a detailed description of the evaluation protocols in Section~\ref{sec:supp-eval}. We describe the baselines we compare to in Section~\ref{sec:supp-baselines}. Finally, in Section~\ref{sec:supp-experiments}, we present additional ablation experiments and more visualizations.
\section{Regularization loss for the generative model}\label{sec:supp-approach}
As mentioned in the main paper, Section.~3.5, in addition to the contrastive and cross-entropy losses, we also use %
an adversarial loss to train the stochastic component of our model. To effectively capture the different modes present in the data and avoid the notorious mode collapse problem of GANs, we apply the recently proposed latent code normalized distance regularizing loss \cite{ndr2, ndr1, yuan2020dlow, zhao2020diverse} defined as
\begin{equation}
   \mathcal{L}_{reg} = - \mathbb{E}_{z_1, z_2} \, \left[ \frac{|| h_v(\mathcal{T}(\mathbf{Q}^{z^1}, \mathbf{M})) - h_v(\mathcal{T}(\mathbf{Q}^{z^2}, \mathbf{M}))||_1}{||z^1 - z^2||_1}\right].
\label{eq:ndr}
\end{equation}

Intuitively, this regularization loss encourages model outputs to be different (in $L_1$-norm), for different input noise vectors $z^1, z^2 \in \mathcal{N}(0, 1)$.
Note that we only apply the regularization loss on the outputs of the visual state branch, \ie $h_v(\cdot)$ in the main paper.
For stronger supervision, in our implementation we follow previous work \cite{ndr1} and generate $S=20$ samples using $S$ latent noise vectors, $\{z^i\}_{1:S}$, calculate the value of $\mathcal{L}_{reg}$ for all possible pairs, and select the maximum value to use in the regularization loss $\mathcal{L}_{reg}$. We validate the role of this regularization loss in Sec.~\ref{sec:supp-experiments}.

\section{Inference with Viterbi algorithm}\label{sec:sup-viterbi}
To find the optimal procedure plan, we use the Viterbi algorithm \cite{viterbi}, as discussed in Section 3.5 of the main paper.
Traditionally, the Viterbi algorithm is used to find the most probable path in the first-order Markov model of a dynamic system.
For a sequence of length $T$ and $N_a$ possible states of the system, the Viterbi algorithm relies on two main inputs: (i) a transition matrix, $A \in \mathbb{R}^{N_a \times N_a}$, capturing the probability of transitioning from one state, $a_i$, to another, $a_j$, (ii) an emission matrix, $B \in \mathbb{R}^{T \times N_a}$, describing the probability of each state, $a_i$, given a set of observations.
In our work, we use the marginal state probabilities for each timestep (defined as $\bar{\Pi}$ in Section 3.5 of the main paper) as the emission matrix, $B$, and estimate the transition matrix, $A$, directly from the ground truth plans.
Specifically, the transition matrix, $A$, is calculated based on the action co-occurence frequencies in the training data. To calculate the value of $A_{i,j}$, we must find the number of times action $a_i$ is followed by action $a_j$ in the ground truth plans. We then normalize each row of $A$ to sum to $1$, by applying $L_1$-normalization followed by a softmax with temperature $\tau=1$.

\section{Implementation details}\label{sec:impl}
Our model uses pre-extracted language and vision features, with a model trained on the HowTo100M dataset for joint text-video embedding \cite{miech19howto100m}. This backbone model embeds both vision and language inputs into $512$ dimensional features. For our model, we use a transformer decoder \cite{vaswani2017attention} with eight heads, two layers and 128 dimensional hidden states. Note that we provide an ablation on our architecture choices in Sec.~\ref{sec:supp-experiments}. Since our pre-trained features are of dimension $512$, we embed our initial features using a multilayer perceptron (MLP) with shape $[512 \rightarrow 256 \rightarrow 128]$ interspersed with ReLU nonlinearities. A similar MLP is used to project the ground truth language features, $l_i$, to the same dimension. For the memory module, we empirically set the number of memory entries, $n$, to $128$. For the critic, $C$, we use another  three layer MLP with shape $[256 \rightarrow 64 \rightarrow 32]$ with ReLU activations in all layers. %
The dimension of the noise vector, $\mathit{z}$, is empirically set to $32$. We train our model for $200$ epochs with an initial learning rate set to $7 \times 10^{-4}$ and decayed by $0.65$ every $40$ epochs. The best performing model on the validation set (\ie randomly collected using 20\% training data as mentioned in main paper Section~4.1) is used to report the final test set results.
\section{Data curation and evaluation protocol}\label{sec:supp-eval}
To train and evaluate our model, we use instructional video datasets to construct sequences with plans of various time horizons, $T$.
Each plan contains a pair of \{start, goal\} visual observations, $\{ v_{start}, v_{goal}\}$, and a sequence of ground truth intermediate action labels, $a_{1:T}$.\\

\noindent \textbf{Data curation.} 
Following previous work \cite{procedure2020, bi2021procedure}, for each instructional video, we start by extracting the \emph{ordered} list of all actions, $a_{1:N}$, present in the video. We also collect language descriptions, $l_{1:N}$, corresponding to each action, $a_i$. In addition, to obtain various start and goal observations, we also extract corresponding visual observations, $v_{1:N}$, by locating the start and end times of each action following previous work \cite{procedure2020, bi2021procedure}. \textit{Note that this information is only used for data curation and not for training as done in previous work}. Each video is therefore described as the ordered set of tuples

\begin{equation}
    V = [(a_1,l_1,v_1), \dotsc, (a_{N},l_{N},v_{N})],
\label{eq:curation}
\end{equation}

\noindent where $N$ is the total number of actions present in the video.

Given the video, $V$, represented with \eqref{eq:curation}, we curate a set of plans with prediction time horizon $T$ by sliding a window of size $T+1$, such that $a_{t+1:t+T}$ and $l_{t+1:t+T}$ represent the set of action plans we need to predict, whereas $v_{t}$ and $v_{t+T}$ represent $v_{start}$ and $v_{goal}$, respectively. \\%This sliding window based data curation procedure is used what we refer to as ``Protocol 1'' in the main paper and is used in all our experiments, unless specified otherwise. \\

\noindent \textbf{Evaluation protocols.} As mentioned in Section 4.3 of the main paper, there exist two different evaluation protocols on CrossTask, namely ``Protocol 1'' and ``Protocol 2''.
In all our experiments, we follow previous work~\cite{procedure2020, bi2021procedure} and use ``Protocol 1'', which relies on the sliding window based data curation procedure described above. In addition, ``Protocol 1'', uses a 70/30 train/test dataset split as mentioned in Section 4.1 of the main paper. For the long horizon planning experiments on CrossTask (\ie Section 4.3 of the main paper), we also adopt an additional protocol, proposed in recent work~\cite{sun2021plate} and referred to as ``Protocol 2''. The main differences of ``Protocol 2'' can be summarized in the following three points: (i) Instead of relying on a sliding window to consider all procedure plans of time horizon $T$ for each video, ``Protocol 2'' randomly selects one procedure plan of horizon $T$ per video. (ii)  ``Protocol 2'' uses 85/15 train/test split. (iii) ``Protocol 2'' predicts actions, $a_{1:T-1}$, which concretely means that, for a prediction horizon, $T$, it actually makes $T-1$ predictions.

\section{Baselines}\label{sec:supp-baselines}
Here, we provide a more detailed description of the procedure planning baselines used in our paper.

\noindent \textit{- Random.} As an elementary baseline, we randomly select action steps from the entire vocabulary with equal probability (\ie Uniform distribution) for evaluation.

\noindent \textit{- Retrieval-Based.} For each start and goal observation pairing, $\{v_{start}, v_{goal}\}$, in the test set, this method retrieves the nearest neighbor in the train-set based on visual feature similarity. The plan labels associated with the retrieved nearest neighbor is used for evaluation.

\noindent \textit{- WLTDO} \cite{ehsani2018let} and \textit{UAAA} \cite{abu2019uncertainty}. These two frameworks use recurrent neural networks (RNN) \cite{rnn} for action planning from the visual observation input.

\noindent \textit{- Universal Planning Networks (UPN)} \cite{srinivas2018universal}. UPN is a physical-world path planning algorithm with a continuous action policy. We follow previous work \cite{procedure2020} to modify it via appending a softmax function for the discrete action space.

\noindent \textit{- Dual Dynamics Networks (DDN)} \cite{procedure2020}. DDN models the state-action transition of procedural plans with a two-branch auto-regressive model. The state branch is modeled with an MLP, while the action branch uses an RNN.

\noindent \textit{- PlaTe} \cite{sun2021plate}. This method simply replaces the models used in the two-branch model of DDN with autoregressive transformer modules.

\noindent \textit{- Ext-GAIL} \cite{bi2021procedure}. This method uses reinforcement learning techniques for procedure planning and augments the DDN architecture with a context latent variable. %

\section{Additional ablation experiments}\label{sec:supp-experiments}
In addition to the various ablations presented in the main paper, here, we provide additional experiments to evaluate other aspects of our loss,  architecture and inference procedure.

\subsection{Impact of the regularization loss}
In this experiment we further evaluate our model's capability to model plan distributions. For this purpose we follow the evaluation procedure adopted in Section 4.4 of the main paper. The effectiveness of the diversity regularization loss, ~\eqref{eq:ndr}, for our approach is studied in Table~\ref{tab:abl_ndr}. It is clear that training without the regularization leads to worse NLL and KL-div values. These results confirm the important role of the regularization in the probabilistic model, which indeed seems to suffer from mode dropping in the absence of such regularization.
\begin{table}[t]
\centering
\resizebox{\columnwidth}{!}{  
\begin{tabular}{l l c c c c}
\toprule %
Metric $\downarrow$ & Method & $T=3$ & $T=4$ & $T=5$ & $T=6$ \\ %
\midrule %
\multirow{2}*{NLL}
& Ours - prob. w/o $\mathcal{L}_{reg}$ & 5.07 & 6.61 & 7.57 & 8.28 \\ %
& Ours - prob. & \textbf{4.89} & \textbf{5.48} & \textbf{6.24} & \textbf{7.67} \\ %
\midrule %
\multirow{2}*{KL-Div}
& Ours - prob. w/o $\mathcal{L}_{reg}$ & 2.37 & 4.45 & 6.49 & 7.74 \\ %
& Ours - prob. & \textbf{2.11} & \textbf{3.50} & \textbf{4.26} & \textbf{6.89} \\ %
\bottomrule %
\end{tabular}
}
\caption{The effect of diversity regularization loss on the produced plans in the CrossTask dataset.}%
\label{tab:abl_ndr}
\vspace{5pt}
\end{table}

\subsection{Impact of the Viterbi post-processing}
In the main paper, we have shown that using Viterbi post-processing, while relying on our predicted plan distribution, $\bar{\Pi}$, as an emission matrix, leads to overall better plan predictions (\ie see Table 1 of the main paper). 
Here, we further study the role of emission, $B$, vs. transition, $A$, matrices, in the Viterbi-based plan inference.
In particular, to show the importance of using our predicted distribution, $\bar{\Pi}$, as an emission matrix, we substitute it with a uniform matrix, \ie a matrix with all values equal to $\frac{1}{N_a}$, and perform plan inference with a correspondingly modified Viterbi algorithm. In other words, in these settings, the plan inference will be driven purely by the transition matrix, $A$. We also experiment with setting $A$ to the uniform matrix, and only using $B$ to drive Viterbi's inference.
Table~\ref{tab:supp-viterbi} shows that our original formulation gives the best results and highlights the importance of estimating the emission and transition matrices properly from data, as handled by our model.
\begin{table}[t]
	\centering
	\resizebox{0.8\columnwidth}{!}{
	\begin{tabular}{c c c c}
		\toprule %
		 emission matrix & \multicolumn{1}{c}{SR $\uparrow$} & \multicolumn{1}{c}{mAcc $\uparrow$} & \multicolumn{1}{c}{mIoU $\uparrow$} \\ %
		\midrule %
		\text{B Uniform} & 0.12 & 7.25 & $<$0.01 \\ %
		\text{A Uniform} & 22.61 & 46.12 &  70.24 \\ %
		\text{Ours} & \textbf{23.34} & \textbf{49.96} & \textbf{73.89} \\ %
		\bottomrule %
	\end{tabular}}
	\caption{
	The effect of estimating transition matrix, $A$, and emission matrix, $B$, of \textit{Viterbi} post-processing from data. The results are for prediction horizon, $T=3$, on the CrossTask dataset.
	}
	\label{tab:supp-viterbi}
\end{table}

\subsection{Impact of the size of the sample set}
In the main paper, we provide inference results obtained by
sampling $\mathcal{K} = 1500$ procedure plans from our probabilistic model. Here, we ablate this parameter and show its influence on how well we fit the ground truth plan distribution. Table~\ref{tab:abl_samples} shows that larger value of $\mathcal{K}$ generally leads to better results as it can better approximate the true distribution. However, it should be noted that larger value of $\mathcal{K}$ require more computation. We therefore chose $\mathcal{K} = 1500$ for our experiments as a reasonable trade-off between performance and computation cost.

\begin{table}[t]
\centering
\resizebox{\columnwidth}{!}{  
\begin{tabular}{l l l l l l}
\toprule %
Metric $\downarrow$ & $\mathcal{K}$ & $T=3$ & $T=4$ & $T=5$ & $T=6$ \\ %
\midrule %
\multirow{4}*{NLL}
& 150 & 5.03 $\pm$ 0.21 & 5.85 $\pm$ 0.31 & 6.78 $\pm$ 0.25 & 8.73 $\pm$ 0.43 \\
& 500 & 4.91 $\pm$ 0.11 & 5.71 $\pm$ 0.20 & 6.57 $\pm$ 0.13 & 8.33 $\pm$ 0.27 \\ %
& 1500 & 4.89 $\pm$ 0.10 & 5.48 $\pm$ 0.11 & 6.24 $\pm$ 0.09 & 7.67 $\pm$ 0.18 \\ %
& 2500 & 4.89 $\pm$ 0.04 & 5.45 $\pm$ 0.06 & 6.22 $\pm$ 0.05 & 7.63 $\pm$ 0.08 \\ %
\midrule %
\multirow{4}*{KL-Div}
& 150 & 2.51 $\pm$ 0.31 & 4.20 $\pm$ 0.44 & 4.66 $\pm$ 0.29 & 7.23 $\pm$ 0.14 \\
& 500 & 2.48 $\pm$ 0.24 & 3.58 $\pm$ 0.12 & 4.51 $\pm$ 0.20 & 6.94 $\pm$ 0.10 \\ %
& 1500 & 2.11 $\pm$ 0.10 & 3.50 $\pm$ 0.06 & 4.26 $\pm$ 0.08 & 6.89 $\pm$ 0.03 \\ %
& 2500 & 2.01 $\pm$ 0.06 & 3.27 $\pm$ 0.02 & 4.18 $\pm$ 0.02 &  6.83 $\pm$ 0.01 \\ %
\bottomrule %
\end{tabular}
}
\caption{Ablation study of the number of samples, $\mathcal{K}$, used by our approach for probabilistic inference. We show NLL and KL-Div results with corresponding variances obtained from 20 runs.}
\label{tab:abl_samples}
\end{table}

\begin{table}[t]
	\centering
	\resizebox{0.7\columnwidth}{!}{
	\begin{tabular}{c c c c}
		\toprule %
		layers/heads & \multicolumn{1}{c}{SR $\uparrow$} & \multicolumn{1}{c}{mAcc $\uparrow$} & \multicolumn{1}{c}{mIoU $\uparrow$} \\ %
		\midrule %
		2/1 & 17.81 & 40.67 & 70.85 \\ %
		2/4 & 20.54 & 48.32 & 73.05 \\ %
		2/8 & \textbf{23.34} & \textbf{49.96} & \textbf{73.89} \\ %
		3/4 & 19.21 & 45.51 & 72.47  \\ %
		3/8 & 14.85 & 39.48 & 69.72 \\ %
		\bottomrule %
	\end{tabular}}
	\caption{
	Ablation study on the number of layers and heads for prediction horizon, $T=3$, with CrossTask.
	}
	\label{tab:abl_arch}
\end{table}

\subsection{Impact of transformer architecture}
Here, we examine the impact of the transformer architecture on the plan prediction performance. To save on the computations, we limit this study to a prediction horizon of $T=3$. Table~\ref{tab:abl_arch} shows the performance of alternative transformer configurations with different number of layers and/or heads. The model used in our paper -- with two layers and eight heads -- performs best on CrossTask.

\subsection{Alternative uncertainty baselines}
Our approach captures the uncertainty of plans by learning a stochastic model (GAN), which allows it to produce distinct procedure plans when seeded with different random noise vectors.
On the other hand, one can follow a more naive approach and obtain diverse outputs from a model trained deterministically (\ie no randomness at training), by adding random noise to the system only at inference, \eg, injecting random noise in the input or using dropout at inference.
Here, we compare our GAN-based formulation to the aformentioned naive approaches for introducing diversity in the predicted 
plans; results are summarized in
Table~\ref{tab:alternative_uncertainty}.
In particular, we evaluate adding Gaussian random noise of different intensity to the input, \ie $\mathcal{N}(0, \sigma)$, and applying dropout (of probability $p$) to the hidden representation of the network at inference time.
We see that both baselines can increase diversity at the cost of accuracy.
The best naive configuration is Ours-deter+dropout ($p=0.3$) which is still inferior to our full GAN-based approach on both deterministic and probabilistic metrics.
This result shows the importance of learning the uncertainty model during training, as opposed to simply adding it at inference, confirming that the GAN is useful for modelling uncertainty in our scenario. 
\begin{table*}[t]
\centering
\resizebox{0.7\textwidth}{!}{  
\begin{tabular}{l c c c c c c c}
\toprule %
Datasets & SR $\uparrow$ & mAcc. $\uparrow$ & mIoU $\uparrow$ & KL-Div $\downarrow$ & NLL $\downarrow$ & MCPrec $\uparrow$ & MCRec $\uparrow$\\ %
\midrule
Ours-prob. (HowTo100M pretrained)  & \textcolor{red}{23.34} & \todo{48.96} &  \todo{73.89} & \todo{2.11} & \textcolor{red}{4.89} & \color{red}{36.61} & \color{red}{66.13} \\ %
Ours-prob. (CrossTask pretrained)   & 10.82 & 35.11 & 58.32 & 3.78 & 5.56 & 17.19 & 12.40\\ %
\midrule
Ours-deter+noise ($\sigma = 1$)  & 20.02 & 44.76 & 72.96 & 2.37 & 5.77 & 24.74 & \todo{67.21}\\ %
Ours-deter+noise ($\sigma = 3$)  & 10.94 & 42.67 & 63.63 & 3.53 & 7.51 & 4.290 & 58.96 \\ %
Ours-deter+noise ($\sigma = 5$)  & 5.83 & 35.53 & 55.43 & 5.52 & 8.53 & 1.169 & 37.01\\ %
Ours-deter+dropout ($p = 0.1$)  & 22.27 & 45.43 & 73.40 & 2.34 & 5.49 & 33.79 & 44.17 \\ %
Ours-deter+dropout ($p = 0.3$)  & 22.04 & 43.64 & 70.97 & 2.46 & 5.83 & 30.61 & 55.32 \\ %
Ours-deter+dropout ($p = 0.5$)   & 20.77 & 44.89 & 70.21 & 2.73 & 6.40 & 24.08 & 64.61 \\ %
\midrule %
Ours-prob. (Strong Sup.)   & \todo{24.41} & \color{red}{45.17} & \color{red}{73.83} & \color{red}{2.12} & \todo{4.71} & \todo{36.89} & 62.69 \\ %
\midrule
\bottomrule %
\end{tabular}
}
\caption{(top rows) Results using features extracted from a model pre-trained on HowTo100M vs.~features finetuned on CrossTask.
(middle rows)  Ablation study on two alternative ways for uncertainty modeling. (last row) Strong visual supervision results. \todo{Blue} text indicates the top performer and \textcolor{red}{red} the $2^{\text{nd}}$-best result. All results are from final evaluation on CrossTask at $T=3$.}
\label{tab:alternative_uncertainty}
\end{table*}

\subsection{Ablation on Video+Language features}
In this study, we choose vision+language features pre-trained on HowTo100M because recent work (e.g., \cite{Luo2020UniVL,miech2020end}) show state-of-the-art results with such features on the datasets we are experimenting with (\ie, COIN, CrossTask). 
Still, it is interesting to consider using a dataset's ``native" features as a variation.
For this experiment, we use the CrossTask dataset, employ the video features trained on it (provided with the dataset) and train language embeddings (not provided by the dataset) by ourselves; we finetune the language model \cite{miech2020end} to align with CrossTask's video features, using contrastive learning.
These features are then used to train our model as described in Section~3 in the main paper. Table~\ref{tab:alternative_uncertainty} (top two rows) shows that using the ``native" CrossTask features is still inferior to the large-scale pre-training on HowTo100M~\cite{miech19howto100m}. This result could be due to the relative scarcity of data in CrossTask that is insufficient to train strong language+vision features.

\subsection{Comparison to strong supervision}
Our framework also can be trained with strong supervision, \ie simply swapping the language supervision signals, $\{l_{i}\}$, in Eq.~5 of the main manuscript with visual ones, $\{v_{i}\}$; see Sec.~\ref{sec:supp-eval} in this supplement. We show the results 
resulting from
this setting in Table~\ref{tab:alternative_uncertainty} (bottom row). Accessing the full annotations can modestly improve or hurt the performance, depending on the metric. Thus, our weakly supervised model performs comparably, while being cheaper to train.

\begin{table}[t]
\centering
\resizebox{\columnwidth}{!}{  
\begin{tabular}{l c c c c c}
\toprule %
Metric $\downarrow$ & Method & $T=3$ & $T=4$ & $T=5$ & $T=6$ \\ %
\midrule
\multirow{2}*{NLL}
& Ours - determinstic & 6.37 & 6.38 & 7.71 & 8.95 \\ %
& Ours - probabilistic & 6.37 & 6.37 &  7.79 & 8.84 \\ %
\midrule
\multirow{2}*{KL-Div}
& Ours - determinstic & 5.34 & 6.88 & 6.75 & 7.06 \\ %
& Ours - probabilistic & 5.75 & 6.68 &  6.84 & 7.15 \\ %
\bottomrule %
\end{tabular}
}
\caption{Evaluation of the plan distributions produced by our probabilistic approach vs. the deterministic variant on COIN.}
\label{tab:prob_coin}
\vspace{5pt}
\end{table}
\begin{table}[t]
\centering
\resizebox{\columnwidth}{!}{  
\begin{tabular}{l c c c c c}
\toprule %
Metric $\downarrow$ & Method & $T=3$ & $T=4$ & $T=5$ & $T=6$ \\ %
\midrule
\multirow{2}*{NLL}
& Ours - determinstic & 7.18 & 7.79 & 8.49 & 9.17 \\ %
& Ours - probabilistic & 7.07 & 7.81 & 8.36 & 9.25 \\ %
\midrule
\multirow{2}*{KL-Div}
& Ours - determinstic & 5.35 & 5.60 & 5.96 & 8.23 \\ %
& Ours - probabilistic & 4.92 & 5.76 & 6.28 & 9.51 \\ %
\bottomrule %
\end{tabular}
}
\caption{Evaluation of the plan distributions produced by our probabilistic approach vs. the deterministic variant on NIV.}
\label{tab:prob_niv}
\vspace{5pt}
\end{table}

\subsection{Additional probabilistic evaluation}
In the main paper, we evaluated our probabilistic model on the CrossTask dataset by measuring the Negative Log-likelihood (NLL) and KL divergence (KL-Div). 
Here, we further provide such results for COIN (in Table~\ref{tab:prob_coin}) and NIV (in Table~\ref{tab:prob_niv}). Interestingly, different from the results obtained on CrossTask (\ie Table 6 in the main paper), we observe that there is no significant difference between our approach and the deterministic counterpart on COIN and NIV. 
We suspect that this happens because the NIV and COIN datasets are lacking variability in goal-conditioned plans.
To verify that hypothesis, we conduct a quantitative evaluation of such variability on COIN and NIV. In Table~\ref{tab:diverse_stats}, we show the average number of distinct plans that can connect the same start and goal observations for each dataset and prediction horizon. It is clear that the plans in CrossTask are much more diverse than those present in the other two datasets. In particular, CrossTask has the largest average number of distinct plans for each time horizon. Notably, for all datasets, it seems longer horizons tend to have larger variability. We believe that this is reasonable because any variations of intermediate steps would result in different paths, which highlights the importance of adopting a probabilistic point of view for procedure planning. 
\begin{table}[t]
\centering
\resizebox{0.8\columnwidth}{!}{  
\begin{tabular}{l c c c c}
\toprule %
Datasets & $T=3$ & $T=4$ & $T=5$ & $T=6$ \\ %
\midrule
CrossTask & 3.26 & 6.76 & 8.40 & 9.29 \\ %
COIN  & 1.51 & 1.93 &  2.25 & 2.35 \\ %
NIV  & 1.03 & 1.07 & 1.28 & 1.29 \\ %
\bottomrule %
\end{tabular}
}
\caption{The average number of unique paths that share the same start and goal across multiple horizons and datasets. }
\label{tab:diverse_stats}
\vspace{5pt}
\end{table}

\subsection{Additional visualizations}\label{sec:supp-visualization}
\begin{figure*}
    \centering
    \resizebox{0.6\linewidth}{!}{
    \includegraphics{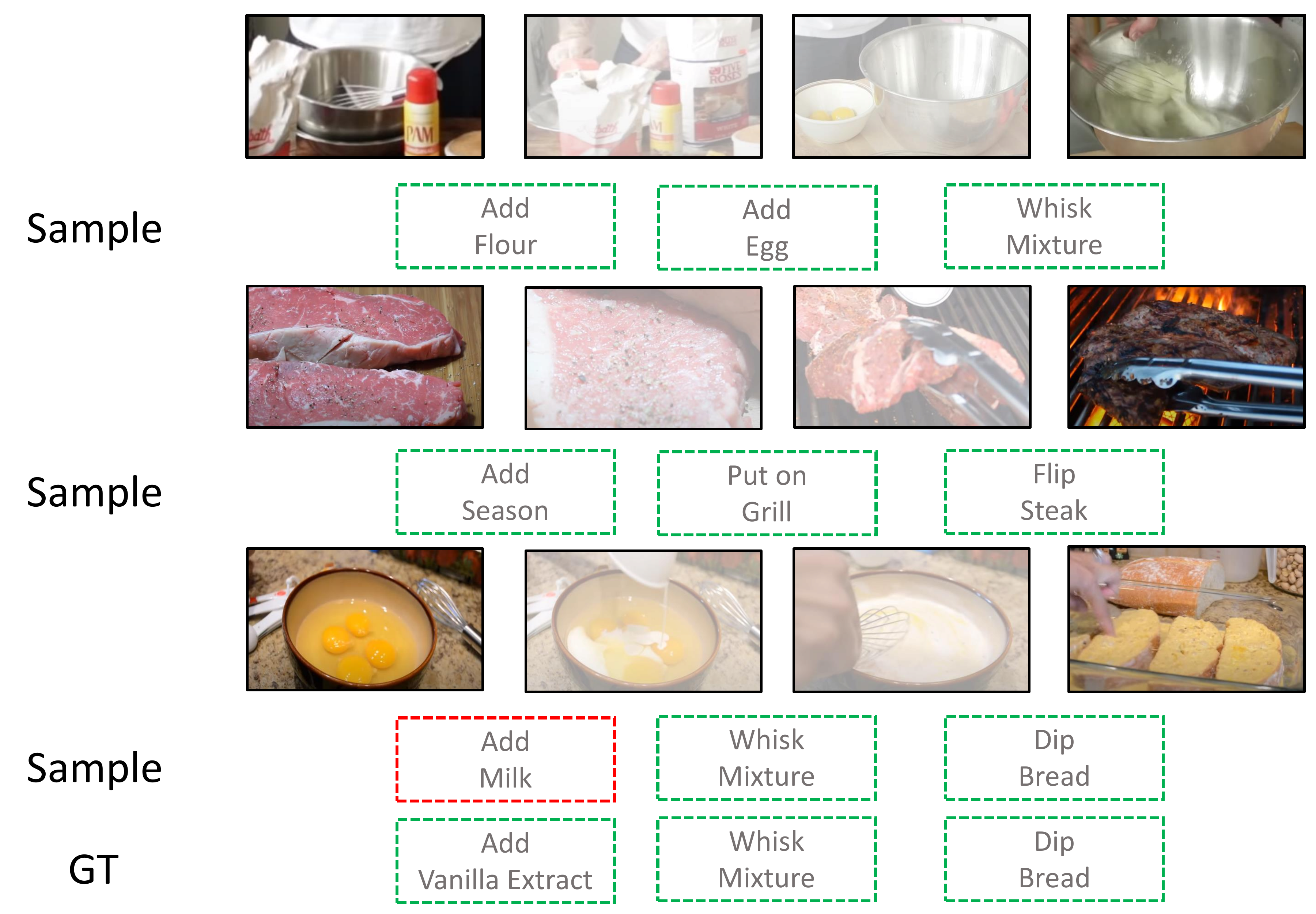}
    }
    \caption{Visualization of two successful outputs (top two rows) and one failure output for $T=3$ (bottom row) on CrossTask. Our approach can produce decent plans when the given start and goal observations are clear. However, we do notice some failures when the steps deviate from the usual expected actions for a certain plan, as shown in the bottom row.
    }
    \label{fig:vis_t3}
\end{figure*}

\begin{figure*}
    \centering
        \resizebox{0.6\linewidth}{!}{

    \includegraphics{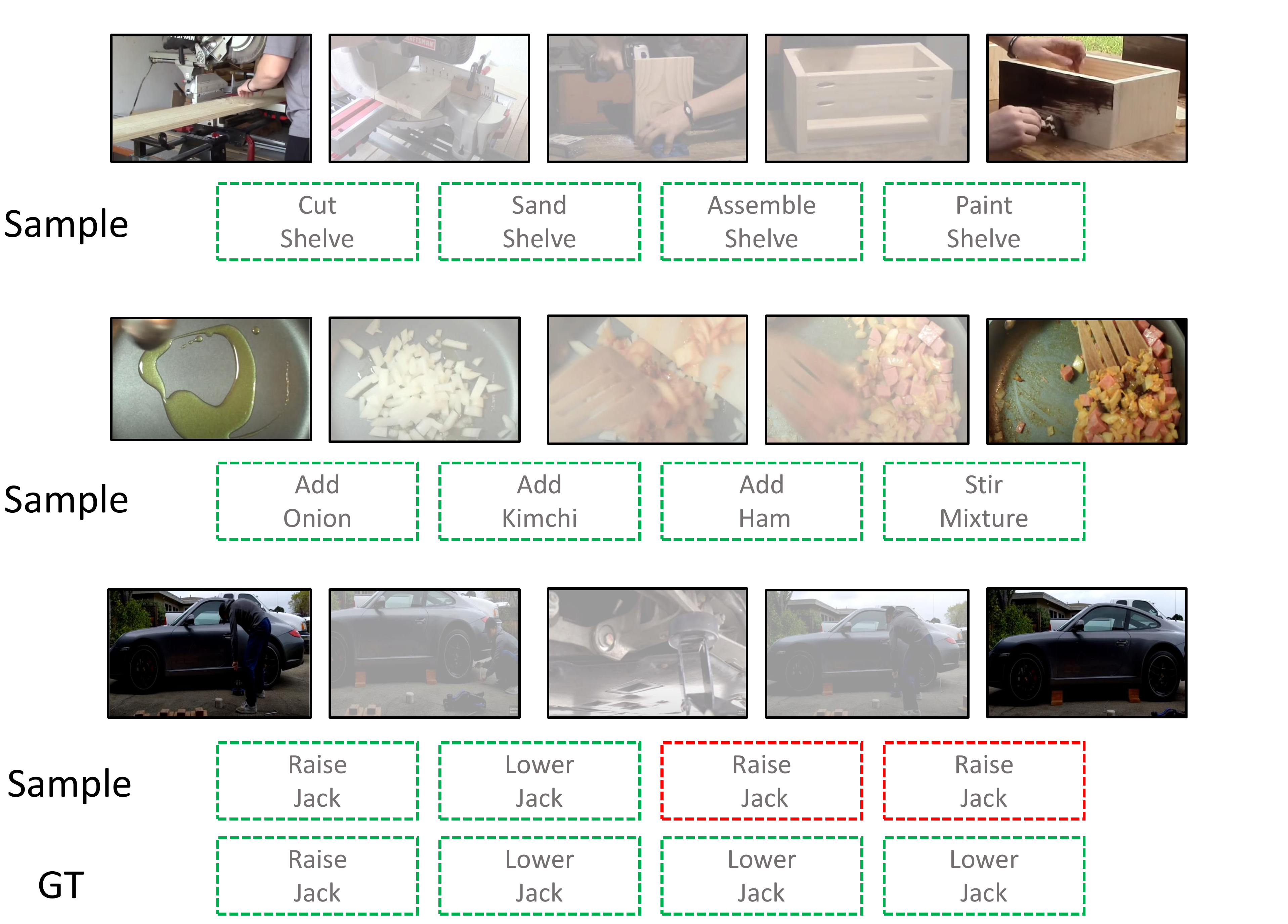}
    }
    \caption{Visualization of two successful outputs (top two rows) and one failure output for $T=4$ (bottom row) on CrossTask. Note that in the failure case, the start and goal observations are not clearly distinguishable. For example, in the procedure of \textit{Jack a Car} shown in the bottom row, it is hard to reason about the transitions between the start and goal observations when their difference is so subtle.}%
    \label{fig:vis_t4}
\end{figure*}

\begin{figure*}
    \centering
        \resizebox{0.7\linewidth}{!}{

    \includegraphics{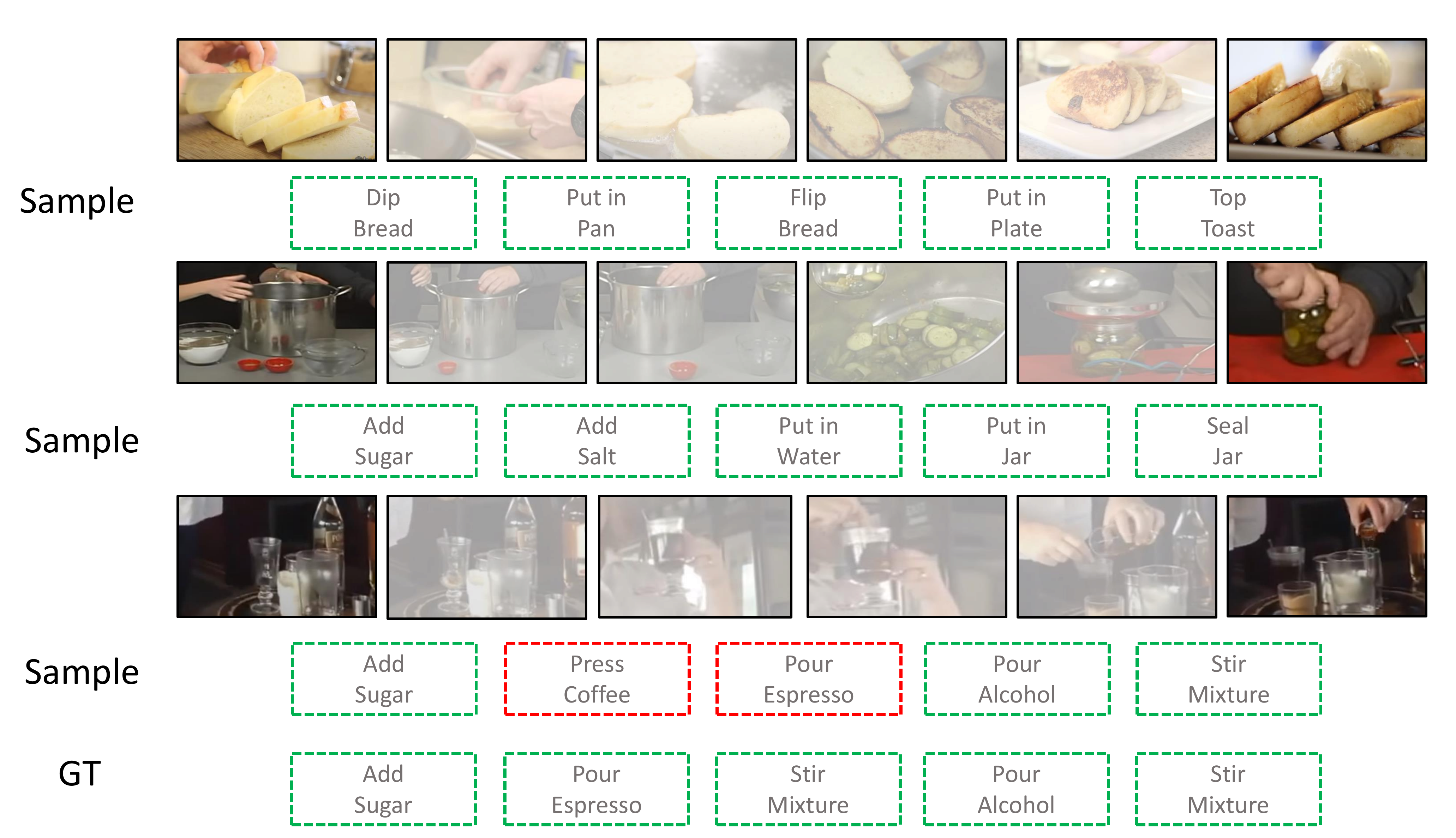}
    }
    \caption{
    Visualization of two successful outputs (top two rows) and one failure output for $T=5$ (bottom row) on CrossTask. In the failure case depicted here, we notice that our model still produces a plausible plan.}%
    \label{fig:vis_t5}
\end{figure*}

\begin{figure*}
    \centering
        \resizebox{0.8\linewidth}{!}{
    \includegraphics{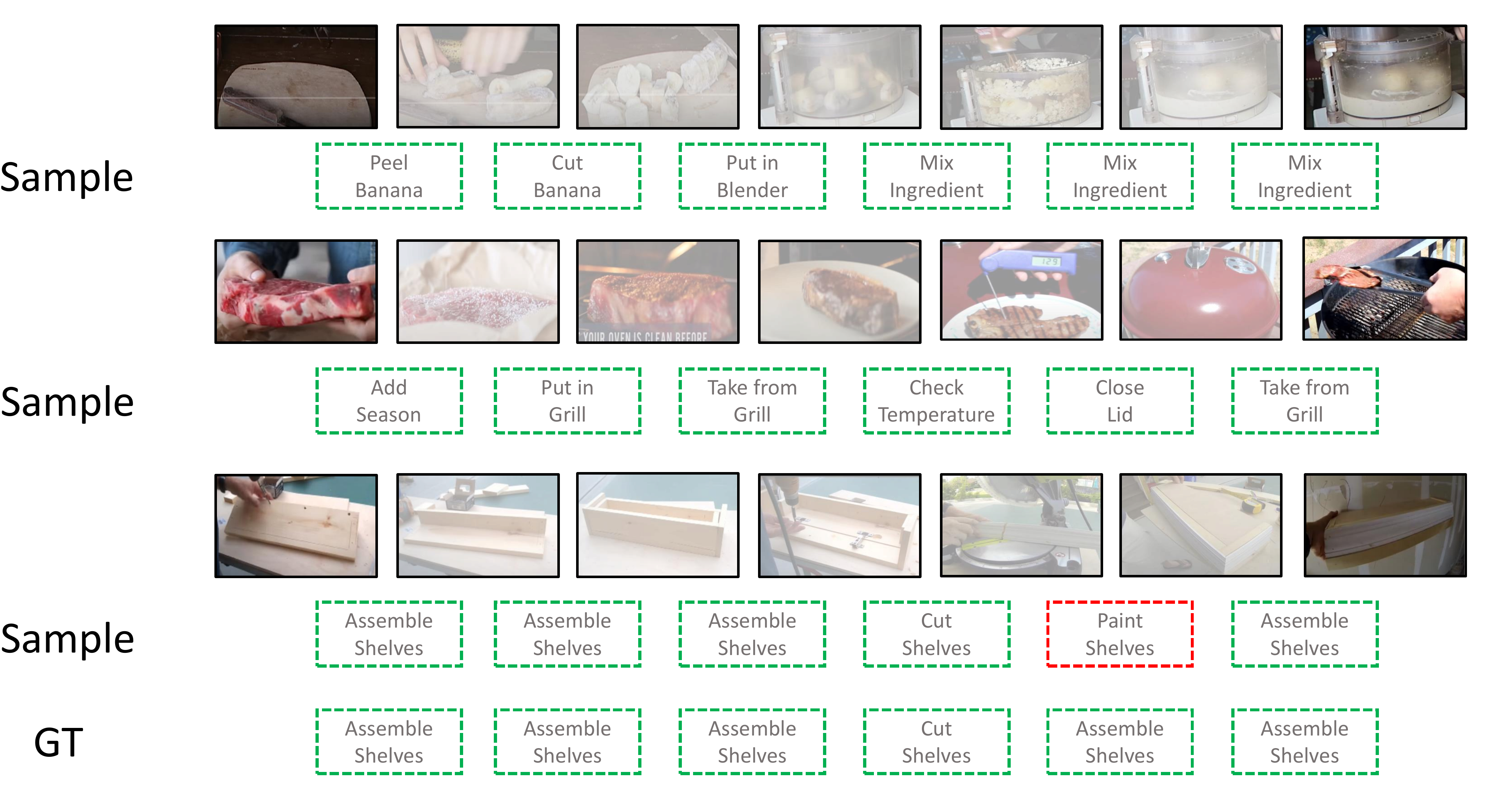}
    }
    \caption{
    Visualization of two successful outputs (top two rows) and one failure output for $T=6$ (bottom row) on CrossTask. Note that the ground truth of the failure case depicted here contains a sequence of repetitive actions. Importantly, notice that while the goal observation depicts a change in color, suggesting the action of \textit{Paint Shelves} as predicted by our model, the ground truth seems to ignore that matter.}%
    \label{fig:vis_t6}
\end{figure*}

In Figures~\ref{fig:vis_t3} to~\ref{fig:vis_t6}, we provide additional visual examples of plans produced by our model for different prediction horizons. We also show failure cases and discuss potential reasons for such behaviour in corresponding captions. Note that the top two rows in each figure are successful predictions and the bottom row is the failure case. In each row of images, the first and last images denote the start and goal observations respectively, and the very next row shows the action labels predicted from our approach (\ie rows beginning with ``Sample''). For failure cases, we show the corresponding ground truth plan  (\ie rows beginning with ``GT'') for better understanding. In addition, we also show the (unseen) intermediate visual observations just for clarity of presentation. Our approach does not use them for training and/or testing.

\section{Attribution of assets}
This research was made possible thanks to the following assets.

\noindent - CrossTask dataset~\cite{CrossTask}. Our study follows the rules from the official license and uses the video URLs and annotations from this website: \url{https://github.com/DmZhukov/CrossTask}. Our usage of CrossTask is limited to this academic work. %

\noindent - COIN dataset~\cite{COIN}. We have signed and submitted the official licence agreement from URL: \url{https://coin-dataset.github.io/}. Our study uses the provided videos/annotations and follows the rules from their license.%

\noindent - NIV dataset~\cite{alayrac2016unsupervised}. We follow the license agreement and use videos/annotations from URL:  \url{https://github.com/jalayrac/instructionVideos}.

\noindent - MIL-NCE pre-trained model~\cite{miech19howto100m}. We follow the license agreement and use the provided pre-trained model weights from URL: \url{https://github.com/antoine77340/S3D_HowTo100M}.

\end{document}